\documentclass[sigconf]{acmart}
 
\copyrightyear{2024}
\acmYear{2024}
\setcopyright{acmlicensed}
\acmConference[CIKM '24] {Proceedings of the 33rd ACM International Conference on Information and Knowledge Management}{October 21--25, 2024}{Boise, ID, USA.}
\acmBooktitle{Proceedings of the 33rd ACM International Conference on Information and Knowledge Management (CIKM '24), October 21--25, 2024, Boise, ID, USA}
\acmISBN{979-8-4007-0436-9/24/10}
\acmDOI{10.1145/3627673.3679669}

\usepackage{microtype}
\usepackage{graphicx}
\usepackage{subfigure}
\usepackage{booktabs} 
\usepackage{hyperref}

\usepackage{amsmath}
\usepackage{amssymb}
\usepackage{mathtools}
\usepackage{amsthm}
\usepackage{url}            
\usepackage{amsfonts}       
\usepackage{nicefrac}       
\usepackage{microtype}      
\usepackage{xcolor}         
\usepackage{graphicx}
\usepackage{amssymb}
\usepackage{bm}
\usepackage{enumitem}
\usepackage{tabularx}
\usepackage{multirow}
\usepackage{caption}
\usepackage{mathtools}
\theoremstyle{plain}
\newtheorem{theorem}{Theorem}[section]
\newtheorem{proposition}[theorem]{Proposition}

\theoremstyle{definition}
\newtheorem{definition}[theorem]{Definition}

\theoremstyle{remark}

\newcommand{\eat}[1]{}
\usepackage[textsize=tiny]{todonotes}
\usepackage[capitalize,noabbrev]{cleveref}

\begin{document}

\title{Data Imputation from the Perspective of Graph Dirichlet Energy}

\author{Weiqi Zhang}
\authornote{Equal Contribution.}
\affiliation{%
  \institution{The Hong Kong University of Science and Technology}
  \city{Hong Kong SAR}
  \country{China}
  }
\email{wzhangcd@connect.ust.hk}

\author{Guanlue Li}
\authornotemark[1]
\affiliation{%
  \institution{The Hong Kong University of Science and Technology (Guangzhou)}
  \city{Guangzhou}
  \country{China}
  }
\email{guanlueli@gmail.com}

\author{Jianheng Tang}
\affiliation{%
  \institution{The Hong Kong University of Science and Technology}
  \city{Hong Kong SAR}
  \country{China}
  }
\email{jtangbf@connect.ust.hk}

\author{Jia Li}
\authornote{Corresponding Author.}
\affiliation{%
  \institution{The Hong Kong University of Science and Technology (Guangzhou)}
  \city{Guangzhou}
  \country{China}
  }
\email{jialee@ust.hk}

\author{Fugee Tsung}
\affiliation{%
  \institution{The Hong Kong University of Science and Technology (Guangzhou)}
  \city{Guangzhou}
  \country{China}
  }
\email{season@ust.hk}

 \begin{abstract} Data imputation is a crucial task due to the widespread occurrence of missing data. Many methods adopt a two-step approach: initially crafting a preliminary imputation (the "draft") and then refining it to produce the final missing data imputation result, commonly referred to as "draft-then-refine". In our study, we examine this prevalent strategy through the lens of graph Dirichlet energy. We observe that a basic "draft" imputation tends to decrease the Dirichlet energy. Therefore, a subsequent "refine" step is necessary to restore the overall energy balance. Existing refinement techniques, such as the Graph Convolutional Network (GCN), often result in further energy reduction. To address this, we introduce a new framework, the Graph Laplacian Pyramid Network (GLPN). GLPN incorporates a U-shaped autoencoder and residual networks to capture both global and local details effectively. Through extensive experiments on multiple real-world datasets, GLPN consistently outperforms state-of-the-art methods across three different missing data mechanisms. The code is available at \url{https://github.com/liguanlue/GLPN}. \end{abstract}





\maketitle


\section{Introduction}\label{sec:introduction}
Missing data is ubiquitous and has been regarded as a common problem encountered by machine learning practitioners in real-world applications \cite{bertsimas2017predictive, le2021sa, kyono2021miracle, lakshminarayan1999imputation, gao2023handling}. In industries, for example, faulty sensors may cause errors or missingness during data collection, which bring challenges to state prediction and anomaly detection \cite{bechny2021missing}. 
 In previous works, a popular paradigm for missing data imputation is the ``draft-then-refine'' procedure \cite{kyono2021miracle, van2011mice, kim2004reuse, bertsimas2017predictive, van2007multiple, li2021deconvolutional, taguchi2021graph, wu2021inductive, mistler2017comparison, wu2023differentiable}. As the name suggests, it first uses basic methods (e.g., mean) to perform a draft imputation of the missing attributes. Then, the completed feature matrix is fed to more sophisticated methods (e.g., graph neural networks  \cite{li2021deconvolutional}) for refinement. However, there is a lack of principle to guide the design of such “draft-then-refine”  procedures: Under what conditions are these two steps complementary? Which aspects of the imputation are important?

In this work, we initialize the first study to examine such a ``draft-then-refine” paradigm from the perspective of Dirichlet energy \cite{cai2020note}, in which it measures the ``smoothness” among different observations. Surprisingly, we find that some rudimentary ``draft” steps of the paradigm, e.g., mean and KNN, will lead to a notable reduction of Dirichlet energy. In this vein, an energy-maintenance ``refine'' step is in urgent need to make the two steps complementary and recover the overall energy. However, we find that many popular methods in ``refine'' step also lead to a diminishment of Dirichlet energy.
For example, Graph Convolution Networks (GCNs) \cite{kipf2016semi, li2017diffusion} have been widely used to serve as a ``refine'' step \cite{wu2021inductive, taguchi2021graph}. Nevertheless, GCN based models suffer from over-smoothness and lead to rapid Dirichlet energy decline \cite{huang2020tackling, min2020scattering, cai2020note}. Considering the above property of GCNs, applying GCN-based models during the ``refine" step will cause further Dirichlet energy diminishment. It means the final imputation may be relatively over-smooth, which is far from accurate imputation from the perspective of energy maintenance.  

Another challenge of the missing data imputation is the shift in global and local distribution in data matrices. Global methods \cite{troyanskaya2001missing,SportisseBJ20,audigier2016multiple} prioritize generating data that approximates the distribution of the original data as a whole, but they may fail to capture the detailed representation of local patterns such as the boundary between clusters. In contrast, local methods \cite{bose2013modified,keerin2021improved,abs-2210-07606,de2016missing} rely on the local similarity pattern to estimate missing values by aggregating information from neighboring data points. However, they may underperform because they disregard global information. 

Regarding the concerns above, we propose a novel deep graph missing data imputation framework, Graph Laplacian Pyramid Network (GLPN). We utilize the graph structure to handle the underlying relational information, where the incomplete data matrix represents node features \cite{rossi2021unreasonable}. The proposed GLPN, consisting of a U-shaped autoencoder and learnable residual networks, has been proven to maintain the Dirichlet energy under the ``draft-then-refine'' paradigm. Specifically, the pyramid autoencoder \cite{lee2019self} learns hierarchical representations and restores the global (low-frequency) information. Based on graph deconvolutional networks \cite{li2021deconvolutional}, the residual network focuses on recovering local (high-frequency) information on the graph. The experiment results on heterophilous and homophilous graphs show that GLPN consistently captures both low- and high-frequency information leading to better imputation performance compared with the existing methods.

To sum up, the contributions of our work are as follows:
\begin{itemize}
\item We analyze general missing data imputation methods with a ``draft-then-refine'' paradigm from the perspective of Dirichlet energy, which could be connected with the quality of imputation.

\item Considering the Dirichlet energy diminishment of GCN based methods, we propose a novel Graph Laplacian Pyramid Network (GLPN) to preserve energy and improve imputation performance at the same time. 

\item We conduct experiments on three different categories of datasets, including continuous sensor datasets, single-graph datasets comprising heterophilous and homophilous graphs, and multi-graph datasets. We evaluated our proposed model on three different types of missing data mechanisms, namely MCAR, MAR, and MNAR.

\item The results demonstrate that the GLPN outperforms the state-of-the-art methods, exhibiting remarkable robustness against varying missing ratios. Additionally, our model displays a superior ability to maintain Dirichlet energy, further substantiating its effectiveness in imputing missing data.

\end{itemize}

The remainder of this paper is organized as follows. Section \ref{sec:related} introduces the related work of imputation methods, the laplacian pyramid and Graph U-Net. Section \ref{sec:prelim} illustrates the task definition and Graph Dirichlet Energy. Meanwhile, a proposition has been introduced to analyze the energy changing of the ``draft-then-refine" missing data imputation paradigm. Section \ref{sec:model} presents the details of the proposed model and Section \ref{sec:analysis} provides the analysis of energy maintenance. In Section \ref{sec:experiment}, we evaluate our method on several real-world datasets to show the superiority of our proposed model. Finally, we draw our conclusion in Section \ref{sec:conclusion}.

\begin{figure*}[tb]
  \centering
  \includegraphics[width=0.95\textwidth]{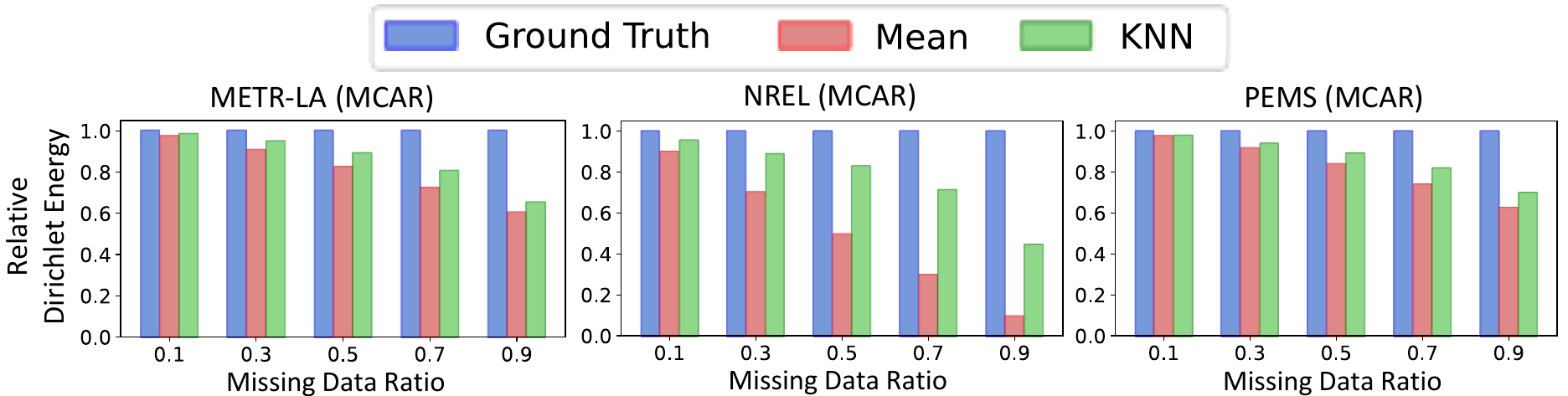}
  \vspace{-4mm}
  \caption{The Dirichlet energy of different imputation strategies on three experimental datasets (METR-LA, PEMS, NREL) with different missing ratios under missing-completely-at-random (MCAR) mechanism. Similar results can be obtained under other mechanisms. Relative Dirichlet energy is normalized by that of the ground truth features.}
\vspace{-4mm}
  \label{fig:energy_change}
  
\end{figure*}

\section{Related Work}
\label{sec:related}

\subsection{Data Imputation}
Classical imputation methods \cite{muzellec2020missing} can be divided into generative and discriminative models. Generative models include the Expectation Maximization (EM) \cite{garcia2010pattern, dempster1977maximum}, Autoencoder \cite{gondara2017multiple}, variational autoenncoders \cite{mattei2019miwae, ivanov2019variational}, and Generative Adversarial Nets (GANs) \cite{yoon2018gain}. Discriminative methods include matrix completion \cite{cai2010singular, hastie2015matrix, troyanskaya2001missing,mazumder2010spectral}, multivariate imputation by chained equations \cite{van2011mice}, optimal transport based distribution matching \cite{muzellec2020missing}, iterative random forests \cite{stekhoven2012missforest}, and causally-aware imputation \cite{kyono2021miracle}. However, both two types overlook the relationship between different observations, i.e., additional graph information.

Recently, several works have extended the Graph Neural Networks (GNNs) to make use of this relational information. GRAPE considers observations and features as two types of nodes in bipartite networks \cite{you2020handling}. GDN proposes to use graph deconvolutional networks to recover input signals as imputation \cite{li2021deconvolutional}. GCNMF adapts GCNs to predict the missing data based on a Gaussian mixture model \cite{taguchi2021graph}. IGNNK utilizes Diffusion GCN to do graph interpolation and estimate the feature of unseen nodes \cite{wu2021inductive}. However, existing GNNs based models tend to give smooth representations for all the observations and lead to an inevitable decrease in graph Dirichlet energy, which will degrade the imputation performance.

Meanwhile, based on the methods used, missing data imputation can also be categorized into two types: global and local approaches. The global approach \cite{EmmanuelMMSMT21, TroyanskayaCSBHTBA01} predicts missing values considering the global correlation information derived from the entire data matrices. However, this assumption may be not adequate where each sample exhibits a dominant local similarity structure. In contrast, the local approach \cite{BoseDGC13} exploits only the local similarity structure to estimate the missing values by computing them from the subset with high correlation. The local imputation technique may perform less accurately than the global approach in homophilous data. To address this limitation, we proposed a hybrid hierarchical method that combines both global and local patterns and is suitable for both homophilous and heterophilous data which could be verified by extensive experiments.

\subsection{Laplacian Pyramid and Graph U-Net}
The laplacian pyramid is first proposed to do hierarchical image compression \cite{burt1987laplacian}. Combining with deep learning framework, deep laplacian pyramid network targets the task of image super-resolution, which reconstructs a high-resolution image from a low-resolution input \cite{lai2017deep, lai2018fast, anwar2020densely}. The main idea of the deep laplacian pyramid network is to learn high-frequency residuals for reconstructing image details, which motivates the residual network design in our proposed model. Graph U-Net \cite{gao2019graph}, consisting of the graph pooling-unpooling operation and a U-shaped encoder-decoder architecture, has been proposed for node classification on graphs. However, Graph U-Net uses skip connection instead of learnable layers as residuals, which would cause over-smoothing node representations and decrease the graph Dirichlet energy, if it is adopted directly for missing data imputation. Besides, some work introduce hierarchical ``Laplacian pyramids" for data imputation \cite{rabin2017missing,rabin2019two}. The Laplacian pyramids in these methods are composed of a series of functions in a multi-scale manner. Differently, the Laplacian pyramid indicates a series of graphs with different scales in our work.

\section{Preliminaries}
\label{pre}
\label{sec:prelim}
\subsection{Task Definition}

Let $\bm X\in \mathbb{R}^{n\times d}$ be a data matrix consisting of $n$ observations with $d$ features for each observation. $\bm X_{i,j}$ denotes the $j^{th}$ feature of $i^{th}$ observation. To describe the missingness in the data matrix, we denote $\bm M\in \{0,1\}^{n\times d}$ as the mask matrix, where $\bm X_{i,j}$ can be observed only if $\bm M_{i,j}=1$. We assume that there exists some side information describing the relationship between different observations, e.g., graph structure. We consider each observation as a node on graph and model the relationship by the adjacency matrix $\bm A\in \mathbb{R}^{n\times n}$.

\noindent \textbf{Task Definition.} Given the observed feature matrix $\bm X^{\star}$, mask matrix $\bm M$, graph structure $\bm A$, the imputation algorithm aims to recover the missingness of data matrix by
\begin{equation*}
    \bm{\hat{X}}=f\left(\bm X^{\star}, \bm A\right),
\end{equation*}
where  $\bm X^{\star}_{i,j}= \begin{cases}\mathrm{NA}, & \text{if } \bm M_{i,j}=0 \\ \bm X_{i,j}, & \text { otherwise }\end{cases}$, $\mathrm{NA}$ represents a missing value, and $f:(\mathbb{R} \cup\{\mathrm{NA}\})^{n\times d} \rightarrow \mathbb{R}^{n\times d}$ denotes a learnable imputation function. 

\subsection{Graph Dirichlet Energy}

Intuitively, the graph Dirichlet energy measures the “smoothness” between different nodes on the graph \cite{cai2020note}. It has been used to measure the expressiveness of the graph embeddings. For data imputation, in the case of reduced graph Dirichlet energy, the recovered missing data suffers from ``over-smoothed" imputation, which leads to poor imputation performance.

\begin{definition}[Graph Dirichlet Energy]\label{de:DE}
Given the node feature matrix $X\in \mathbb{R}^{n\times d}$, the corresponding Dirichlet Energy is defined by:
\begin{equation}
\resizebox{\columnwidth}{!}{$
    E_D(\bm X)=tr\left(\bm X^T \tilde{\Delta} \bm X\right)=\dfrac{1}{2}\sum_{i,j=1}^{n}\bm A_{i,j}  || {\dfrac{\bm X_{i,:}}{\sqrt{1+\bm D_{i,i}}}-\dfrac{\bm X_{j,:}}{\sqrt{1+\bm D_{j,j}}}} ||^2 ,$}
\end{equation}
\end{definition}

\noindent where $tr(\cdot)$ denotes the trace of a matrix and $\parallel\cdot\parallel$ is the $\ell^{2}$-norm. $\bm X_{i,:}$ is the $i^{th}$ row of feature matrix $\bm X$ corresponding to the features of the $i^{th}$ node. $\bm D_{i,i}$ is the $i^{th}$ element on the diagonal of the degree matrix $\bm D$. $\tilde{\Delta}:=\bm I_n-\tilde{\bm D}^{- \frac{1}{2}}\tilde{\bm A}\tilde{\bm D}^{-\frac{1}{2}}$ is the augmented normalized Laplacian matrix, where $\tilde{\bm A}:=\bm A+\bm I_n$ and $\tilde{\bm D}:=\bm D+\bm I_n$ denote the adjacency and degree matrix of the augmented graph with self-loop connections.

The following equation connects the graph Dirichlet energy with the quality of imputation.
\begin{equation}\label{eq:di}
\parallel\hat{\bm X} - \bm X\parallel \geq \frac{|E_D(\hat{\bm X}) - E_D(\bm X)|}{2B\lambda_{\max}},
\end{equation}
where $B = \max(\parallel\hat{\bm X}\parallel, \parallel\bm X\parallel)$, and $\lambda_{\max}$ is the largest eigenvalue of $\tilde{\Delta}$ which is smaller than 2. 
The proof of Equation (\ref{eq:di}) is in the Appendix. 
{It reveals that when the Dirichlet energy gap becomes larger, the lower bound of the $l_2$ distance is larger. In other words, if an imputation method cannot keep the Dirichlet energy, the optimal imputation quality of this method is limited by the Dirichlet energy gap.} Thus, constraining the Dirichlet energy of $\hat{\bm X}$ close to $E_D(\bm X)$ is a necessity for good imputation performance.

\subsection{Draft-then-Refine Imputation}

As discussed before, the ``draft-then-refine'' procedure has no guarantee of maintaining Dirichlet energy. To further justify the insight, we focus on the ``draft'' imputation stage and find that the Dirichlet energy of the ``draft'' imputation matrix $\hat{\bm X}$ is usually less than that of the ground truth matrix $\bm X$. 

On the theoretical side, in the following proposition, we analyze a class of imputation methods where each missing feature is filled by a convex combination of observed features from the same column. 

\begin{proposition}\label{proposition2}
Suppose each element in $\bm X$ is identically independent drawn from a certain distribution whose first and second moment constraints satisfy $\mathbb E(\bm X_{i,j}) = 0$ and $\text{Var}(\bm X_{i,j}) = 1$, and the imputation $\hat{\bm X}$ satisfy $\hat{\bm X}_{i,j}=\sum_{k\in S_k}\alpha_{k}\bm X^{\star}_{i,k}$, where $\sum_{k\in S_k} \alpha_{k}=1$, $\alpha_k \geq 0$, and $S_k = \{1\leq k \leq n|\bm M_{i,k}=1\}$, we have $\mathbb E[E_D(\hat{\bm X})] \leq \mathbb E[E_D(\bm X)]$.
\end{proposition}
We refer the reader to Appendix for the proof.
Proposition \ref{proposition2} generalizes those methods that use mean, KNN, or weighted summation for the ``draft'' imputation and shows the universality of energy reduction. In practice, the distribution of $\bm X$ may not strictly follow the above assumption. Nonetheless, the reduction of Dirichlet energy still exists in the real-world imputation datasets. As shown in Figure \ref{fig:energy_change}, imputation strategies during the ``draft" step will lead to the Dirichlet energy reduction. As the missing ratio increases, the Dirichlet energy of imputation keeps decreasing, which brings more challenges for further refinement.

\begin{figure*}[tb]
  \centering
  \includegraphics[width=\textwidth]{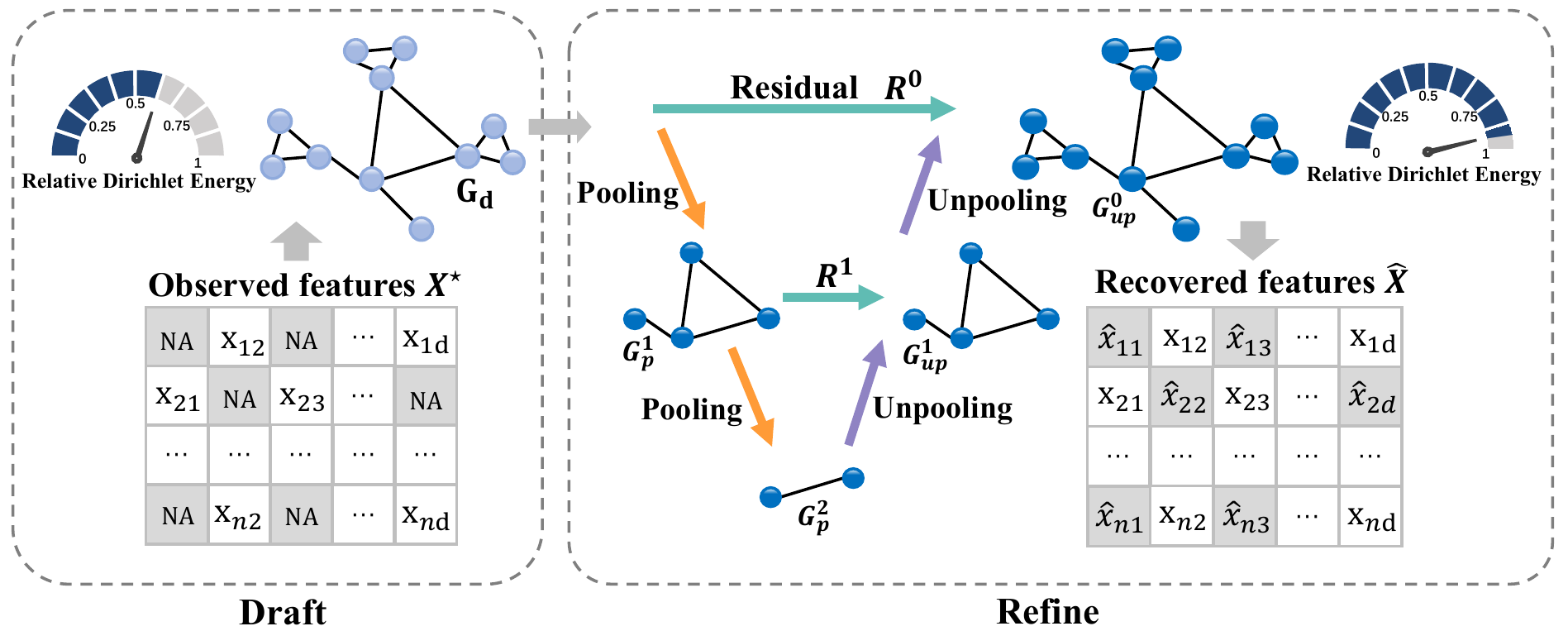}
  \vspace{-6mm}
  \caption{Overview of proposed Graph Laplacian Pyramid Network. The U-shaped autoencoder extracts the clustering and coarse patterns, while the Residual Network reconstructs local details. By combining these two parts, the model can refine the draft imputation and preserve the Dirichlet energy. The dashboards indicate the relative Dirichlet energy of draft and refined features.}
  \vspace{-4mm}
  \label{Graph_Lap}

\end{figure*}

\section{Model Design}
\label{sec:model}

Inspired by the analysis above, we propose Graph Laplacian Pyramid Network (GLPN) to improve imputation performance and reduce the Dirichlet energy reduction during imputation tasks. Following the ``draft-then-refine" procedure, during the first stage, we use a general method to construct the draft imputation of the missing features. During the second stage, GLPN is used to refine the node features while preserving the Dirichlet energy of the graph. 
Figure \ref{Graph_Lap} depicts the whole framework of our model. 

\subsection{Draft Imputation}
The model takes a missing graph $G = (\bm X^{\star}, \bm A)$ as input. During the first stage, we use Diffusion Graph Convolutional Networks (DGCNs) \cite{li2017diffusion} to get the initial imputation of missing node features. The layer-wise propagation rule of DGCNs is expressed as:
\begin{equation}
\resizebox{\columnwidth}{!}{$
\bm H^{(k+1)}=\sigma \left(\sum^{M-1}_{m=0} \left (\theta_{m,1}\left(\bm D^{-1}_O \bm A \right)^m+\theta_{m,2}\left(\bm D^{-1}_I  \bm A^T \right)^m\right) \bm H^{(k)}\right),$}
\end{equation}
where $\bm D_O$ and $\bm D_I$ are the out- and in-degree diagonal matrix. $M$ is the diffusion step. $\sigma$ is the activation function. $\theta \in \mathbb{R}^{M \times 2}$ is the parameters of the filter. $\bm H^{(k)}$ is the feature matrix after $k$-steps DGCNs. With input $\bm H^{(0)}= \bm X^{\star}$, we use $K$-layer DGCNs to get draft imputation $G_d = (\bm X_d,\bm A)$, where $\bm X_d=\bm H^{(K)}$.

\subsection{GLPN Architecture}

During the second stage, GLPN aims to refine the draft features for imputation.
The architecture of GLPN consists of two branches: The \textbf{U-shaped Autoencoder} obtains coarse and hierarchical representations of the graph, while the \textbf{Residual Network} extracts the local details on the graph. 

The U-shaped autoencoder captures the low-frequency component, which will generally lead to a low Dirichlet energy. Thus, to alleviate the Dirichlet energy decline, high-frequency components are expected. In this vein, the residual network acts as the high-pass filter to get the high-frequency component.

As shown in Figure \ref{Graph_Lap}, we construct a 2-level graph pyramid for illustration. $G^l_p = (\bm X^l_p, \bm A_p^l)$ is the reduced and clustered version of $G_d$ after the pooling operator at $l^{th}$ pyramid. $G^l_{up} = (\bm X^l_{up}, \bm A_{up}^l)$ denotes the $l^{th}$ level reconstructed graph after the unpooling operator. $\bm R^i$ is the detailed local information obtained via the residual network to overcome the energy decline issue. At each layer of decoder, $\bm X^l_{up}$ and $\bm R^l$ are combined and fed to the next level.

\subsection{U-shaped Autoencoder}

The U-shaped autoencoder aims at getting the global and clustering pattern of the graph. The encoder uses pooling operators to map the graph into the latent space. The decoder then reconstructs the entire graph from latent space by combining residuals and unpooled features.

\noindent \textbf{Pooling and Unpooling:} 
We use the self attention mechanism \cite{lee2019self, li2019semi} to pool the graph into coarse-grained representations. We compute the soft assignment matrix $\bm S^{(l)}$ as follows:
\begin{equation}
\bm S^{(l)}=\operatorname{softmax}\Big(\operatorname{tanh}\big(\bm X_p^{(l)}\bm W_1^{(l)}\big)\bm W_2^{(l)}\Big),
\end{equation}
where $\bm S^{(l)} \in \mathbb{R}^{n_l \times n_{l+1}}$ donates the assignment matrix at the $l^{th}$ pooling layer, $\bm W_1^{(l)}$ and $\bm W_2^{(l)}$ are two weight matrices of the $l^{th}$ level pyramid. 
Then the coarsen graph structure $\bm A_p^{(l)}$ and features $\bm X_{p}^{(l)}$ at $l^{th}$ pyramid are 
\begin{equation}
\bm X_{p}^{(l+1)}  = {{\bm S^{(l)}}^T}\bm X_p^{(l)}; \\
~ \bm A_p^{(l+1)}  = \operatorname{softmax}\left({\bm S^{(l)}}^T\bm A_p^{(l)}\bm S^{(l)}\right).
\end{equation}

As for the unpooling operator, we just reverse this process and get the original size graph by

\begin{equation}
\resizebox{\columnwidth}{!}{$
\bm X_{up}^{(l-1)}  = \bm S^{(l)} \left(\bm X_{up}^{(l)} \oplus \bm R^{(l)}\right); ~ \bm A_{up}^{(l-1)}=\operatorname{softmax}\left(\bm S^{(l)}\bm A_{up}^{(l)}{\bm S^{(l)}}^T\right),$}
\end{equation}
where $\oplus$ is combination operator. 
In the bottleneck of a $L$-level GLPN, $\bm X_{up}^{(L)}=\bm X_{p}^{(L)}, \bm A_{up}^{(L)}=\bm A_{p}^{(L)}$. 
Given the draft imputation $\bm X_p^{(0)} = \bm X_d, \bm A_p^{(0)}= \bm A$, we can get the final refined feature matrix $\hat{\bm X} = (\bm X_{up}^{(0)} \oplus \bm R^{(0)})$.

\subsection{Residual Network}
The goal of the residual network is to extract the high-frequency component of the graph and maintain the Dirichlet energy. We adopt Graph Deconvolutional Network (GDN) \cite{li2021deconvolutional} which uses inverse filter $g^{-1}_c(\lambda _i)=\frac{1}{1-\lambda_i}$ in spectral-domain to capture the detailed local information. Similar to \cite{li2021deconvolutional}, we use Maclaurin Series to approximate the inverse filter. Combining with wavelet denoising, the $M^{th}$ order polynomials of GDN can be represented as:
\begin{equation}\label{eq:GDN}
\resizebox{0.8\columnwidth}{!}{$
\bm R^{(l)}= \sigma \left( \left(\bm I_n+\sum^M_{m=1}\frac{(-1)^m}{m!}\bm \tilde{\Delta}^m\right)\bm X_p^{(l)} \bm W_3^{(l)} \right), $}
\end{equation}
where $\bm W_3^{(l)}$ is the parameter set to be learned. 

The model is trained to accurately reconstruct each feature from the observed data by minimizing the reconstruction loss:
\begin{equation}
\mathcal{L} =  \parallel\hat{\bm X} - \bm X \parallel ^2    ,
\end{equation}
where $\hat{\bm X}$ and $\bm X$ are the imputed and ground truth feature vectors, respectively.

\section{Energy Maintenance Analysis}
\label{sec:analysis}

Here, we will discuss how GLPN preserves Dirichlet energy compared with vanilla GCN \cite{kipf2016semi}.

\subsection{Graph Convolutional Networks}

 First, we present the Dirichlet energy analysis of GCN. The graph filter of GCN can be represented by $\bm P=\bm I-\tilde{\Delta}$. Each layer of vanilla GCN can be written as:
\begin{equation}
\bm X^{(l+1)} = \sigma(\bm P \bm X^{(l)}\bm W^{(l)}).
\end{equation}
After removing the non-linear activation $\sigma$ and weight matrix $\bm W^{(l)}$, the Dirichlet energy between neighboring layers could be represented by \cite{ zhou2021dirichlet}:
\begin{equation}
    \left(1-\lambda_{1}\right)^{2} E_D(\bm X^{(l)}) \leq E_D(\bm X^{(l+1)})
    \label{eq:pre},
\end{equation}
where $\lambda_1$ is the non-zero eigenvalues of matrix $\tilde{\Delta}$ that is most close to values 1. {We refer to this simplified Graph Convolution \cite{wu2019simplifying} as simple GCN. Here, we only retain the lower bound as we focus on the perspective of energy decline.} Since the eigenvalues locate within the range $[0,2)$, $\left(1-\lambda_{1}\right)^{2}$ may be extremely close to zero, which means the lower bound of Equation (\ref{eq:pre}) can be relaxed to zero. For a draft imputation $\bm X_d$, {multi-layer simple GCN has no guarantee of maintaining  $E_D(\bm {\hat{X}})$ in the refinement stage.}

\subsection{GLPN}

\label{sec:glpn}
Now we analyze the energy maintenance of GLPN. We use the first-order Maclaurin Series approximation of GDN (Equation (\ref{eq:GDN})) to simplify the analysis, which could also be called as Laplacian Sharpening \cite{park2019symmetric} with the graph filter $\bm P_l=\bm I+\tilde{\Delta}$. 

For the proposed GLPN with one-level pooling-unpooling, the output of our model could be formulated by: 
\begin{equation}
    \hat{\bm X}=\bm P_l \bm X_d+ \alpha \bm S\bm S^T\bm X_d,
    \label{eq.pyramid_layer1}
\end{equation}
where $\bm X_d \in \mathbb{R}^{n\times d}$ is the draft imputed feature matrix of graph and $\hat{\bm X}$ is the refined feature. $\bm S\in \mathbb{R}^{n \times n_0} (n>n_0)$ should be the assignment matrix and the sum of row equals to 1 (i.e. $\operatorname{sum}(\bm S_{i:})=1, i \in [1, n])$. The first term $\bm P_l=\bm I+\tilde{\Delta}$ from the residual network is used as a high-pass filter to capture the high-frequency components. 
The second term comes form the U-shaped autoencoder that obtains the global and low-frequency information.

\begin{proposition}
\label{proposition}
{If we use the first-order Maclaurin Series approximation of GDN as the residual layer}, the Dirichlet energy of one layer {simplified} GLPN {without the weight matrix} is bounded as follows: 
\begin{equation} 
(1 + C_{min})^2 E_D(\bm X_d) \leq E_D (\hat{\bm X}),
\end{equation}
\end{proposition}
\noindent where $C_{min}$ is the minimum eigenvalue of matrix $\tilde{\Delta}+\alpha \bm S\bm S^T$. Meanwhile, for multi-layer GLPN, we can obtain a similar result that {simplified} GLPN can \eat{amplify and }maintain the Dirichlet energy compared with simple GCN in Equation (\ref{eq:pre}). 
Thus, with a non-zero bound, GLPN alleviates the decline of Dirichlet energy during the ``refine" stage. We refer the reader to the Appendix for the proof.

\section{Experiments}
\label{sec:experiment}

\label{sec:experiment0}
\begin{figure*}[tb]
  \centering
  \includegraphics[width=\textwidth]{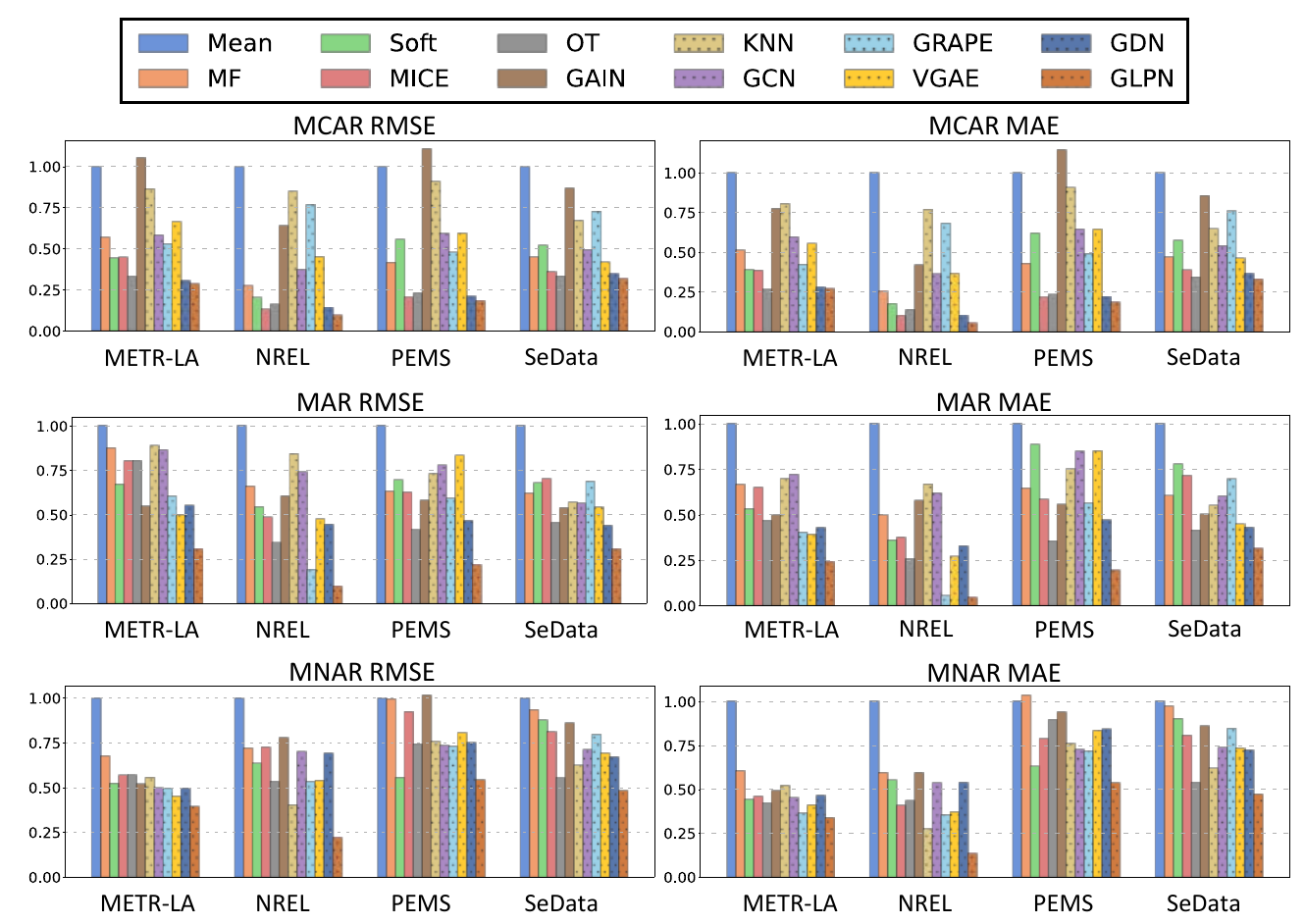}
  \vspace{-7mm}
  \caption{Imputation results on four benchmark datasets with different settings: MCAR (top), MAR (middle), and MNAR (bottom). Both RMSE (left) and MAE (right) are normalized by the performance of mean imputation.}
  \vspace{-4mm}
  \label{fig:Experiment_Result}

\end{figure*}

\subsection{Datasets}

We conduct extensive experiments to evaluate the performance of our proposed framework on three categories of datasets: Continuous sensor datasets, Single-graph graph datasets and Multi-graph datasets. 

\begin{itemize}
\item \textbf{Continuous sensor datasets} \cite{wu2021inductive}, including three traffic speed datasets, METR-LA \cite{li2017diffusion}, PEMS \cite{li2017diffusion}, SeData \cite{cui2018deep} and one solar power dataset, NREL \cite{cui2019traffic}.

\item \textbf{Single-graph datasets} including three heterophilous graphs Texas, Cornell \cite{garcia2016using}, ArXiv-Year \cite{lim2021large} and three homophilous graphs Cora, Citeseer \cite{kipf2016semi}, YelpChi \cite{lim2021large}. 

\item \textbf{Multi-graph datasets} including SYNTHIE \cite{morris2016faster}, PROTEINS \cite{borgwardt2005protein} and FRANKENSTEIN \cite{orsini2015graph}. 

\end{itemize}


\begin{figure}[tb]
  \centering
  \includegraphics[width=\columnwidth]{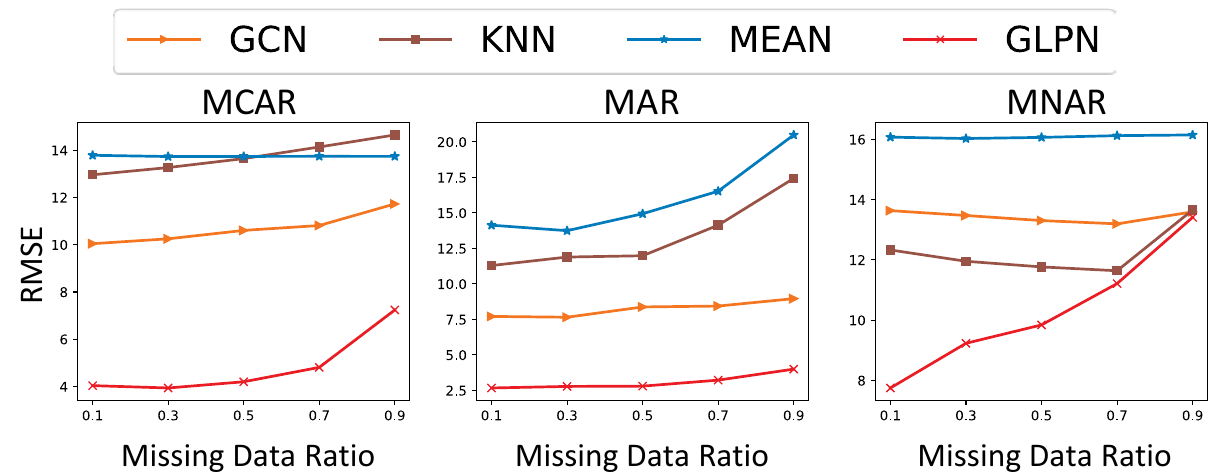}
\vspace{-4mm}
  \caption{RMSE for METR-LA with different data missing ratios. Both structure-free and structure-based baselines are compared under three missing mechanisms.}
  \vspace{-4mm}
  \label{fig:robustness}

\end{figure}

\vspace{-2mm}
\subsection{Baselines}

We compare the performance of our proposed model with the following baselines, which could be further divided into two categories: \textbf{Structure-free methods}: MEAN, MF, Soft \cite{hastie2015matrix}, MICE \cite{van2011mice}, OT \cite{muzellec2020missing}, GAIN \cite{yoon2018gain}; \textbf{Structure-based methods}: KNN, GCN \cite{kipf2016semi}, GRAPE \cite{you2020handling}, VGAE \cite{kipf2016variational}, GDN \cite{li2021deconvolutional}. To be specific, the structure-free methods ignore the graph structure of the data, which leads to a pure matrix imputation task. On the contrary, the structure-based methods take the graph structure into consideration while performing imputation.

\vspace{-2mm}
\subsection{Experimental Setup}

\label{subsec:experimental_setup}
\subsubsection{Tasks and Metrics} The imputation performance will be evaluated by using three common missing data mechanisms: Missing Completely At Random (MCAR), Missing At Random (MAR), and Missing Not At Random (MNAR) \cite{rubin1976inference}. 

MCAR refers to a missing value pattern where data points have vanished independently of other values. We generate the missingness at a rate of 20\% uniformly across all the features. MAR is also a missing pattern in which values in each feature of the dataset are vanished depending on values in another feature.
For MAR, we randomly generate 20\% of features to have missingness caused by another disjoint subset of features. 

In contrast to MCAR and MAR, MNAR means that the probability of being missing depends on some unknown variables \cite{mohan2013graphical}. In our experiments, for MNAR, the features of some observations are entirely missing.
MNAR refers to the cases of virtual sensing or interpolation, which estimates the value of observations with no available data.
We use the Mean Absolute Error (MAE) and Root Mean Squared Error (RMSE) as performance metrics. 

\subsubsection{Parameter Settings}
For all experiments, we use an 80-20 train-test split.
For experimental configurations, we train our proposed model for 800 epochs using the Adam optimizer with a learning rate of 0.001. For the "draft" imputation stage, we use a 1-layer DGCN with 100 hidden units to obtain the rudimentary imputation. We build the residual network using one GDN layer approximated by three-order Maclaurin Series. 
For the pooling-unpooling layer, we set the number of clusters $r$ as 60-180 according to the size of dataset, respectively. 

\begin{table}[tb]
\centering
\caption{The ablation study under MCAR mechanism with 20\% data missing ratio. `w/o R' and `w/o P' mean the variants of GLPN without residual network and U-shaped autoencoder respectively. }
\resizebox{0.85\columnwidth}{!}{
\begin{tabular}{ccccc}
\toprule
\textbf{Variant}             & \textbf{METR-LA} & \textbf{NREL}  & \textbf{PEMS}  & \textbf{SeData} \\ \midrule
\textbf{w/o R} & 5.345            & 1.121          & 2.565          & 5.963           \\
\textbf{w/o P}  & 4.419            & 1.285          & 3.450          & 6.618           \\ \midrule
\textbf{GLPN}         & \textbf{4.079}   & \textbf{0.866} & \textbf{1.546} & \textbf{3.555}  \\ \bottomrule
\end{tabular}
}
\label{ablation}
\vspace{-4mm}
\end{table}

\begin{table*}[tbp]
\centering
\caption{Imputation and classification performance on heterophilous and homophilous graphs (missing ratio = 0.8). ``Full'' and ``Miss'' represent classification results with full and missing features, respectively.}
\vspace{-4mm}
\resizebox{0.9\textwidth}{!}{
\begin{tabular}{c|cccccc|cccccc}
\toprule
               & \multicolumn{6}{c|}{\textbf{Heterophilous graphs}}  & \multicolumn{6}{c}{\textbf{Homophilous graphs}}   \\ \midrule
               &  \multicolumn{2}{c}{\textbf{Texas}}  &  \multicolumn{2}{c}{\textbf{Cornell}}  &  \multicolumn{2}{c|}{\textbf{ArXiv-Year}}  & \multicolumn{2}{c}{\textbf{Citeseer}}  & \multicolumn{2}{c}{\textbf{YelpChi}}  &  \multicolumn{2}{c}{\textbf{Cora}} \\ 
               &  \multicolumn{2}{c}{h = 0.06}  &  \multicolumn{2}{c}{h = 0.11}  &  \multicolumn{2}{c|}{h=0.22}  &  \multicolumn{2}{c}{h=0.74} &  \multicolumn{2}{c}{h=0.77}  & \multicolumn{2}{c}{h = 0.81} \\  
               & MAE & ACC(\%) & MAE & ACC(\%) & MAE & ACC(\%) & MAE & ACC(\%) & MAE & ACC(\%) &  MAE & ACC(\%) \\
               \midrule
\textbf{Full} & - &55.68 &-&55.95  & - &44.60  & - & 59.36 & -&63.63  & - & 77.20  \\
\textbf{Miss}  & - & 49.51& - &51.12& - & 41.17 & - & 43.64& -&61.39 & -  & 62.47  \\  \midrule
\textbf{Mean} &0.0890 & 50.81 & 0.0998 &52.43  & 0.1736 & 41.66 & 0.0166  & 44.07 &0.3772  &60.88  &0.0340 & 62.03 \\
\textbf{KNN}  & 0.1130 &52.97   & 0.0899 &50.97 &0.1637 & 42.86 &0.0165 &  45.98 & 0.3313 & 61.53 &  0.0302 & 62.17\\
\textbf{GRAPE} &0.0935 & 53.86&0.0955 & 52.73& 0.1656& 41.76 &0.0166 & 44.34 &0.2631 & 60.31 &0.0281 & 62.76  \\
\textbf{VGAE} & 0.0954& 53.24& 0.1057& 53.93 & 0.1698& 41.47 & 0.0165&44.17 & 0.2731& 61.07& 0.0351& 60.83  \\
\textbf{GCN}   & 0.0833 & 53.77 & 0.0909 & 53.30  &0.1639& 42.06 &0.0163& 44.12  & 0.2665& 60.75& 0.0275 & 61.50  \\
\textbf{GDN}  & 0.0825&54.06 &0.0885 &54.02 & 0.1595& 42.12& 0.0165&44.40 &0.2717 & 59.62&0.0276& 62.57  \\
\textbf{GLPN}  & \textbf{0.0774}&\textbf{55.49}  & \textbf{0.0868 }& \textbf{54.59 }& \textbf{ 0.1592 }&\textbf{43.18 }    & \textbf{  0.0157 }    &  \textbf{46.06}   & \textbf{0.2502}&\textbf{63.21} & \textbf{     0.0261}  & \textbf{63.34}   \\
\bottomrule
\end{tabular}
}
\label{table:non-sensor1}
\end{table*}


\subsection{Evaluation on Continuous Sensor Datasets}

As shown in Figure \ref{fig:Experiment_Result}, for all the three missing data mechanisms, our proposed GLPN has the best imputation performance with the lowest MAE and the lowest RMSE. Empirical results show that GLPN decreases the overall imputation error by around 9\%.
In general, structure-based methods (denoted by dotted bars) outperform structure-free methods (denoted by filled bars) by taking advantage of graph structure as side information. 
Compared with other structure-based deep learning models, such as GCN based methods, GLPN achieves the greatest performance gain thanks to the design of graph Dirichlet energy maintenance therein.

\subsubsection{Robustness against Different Missing Ratios}

To better understand the robustness of GLPN, we conduct the same experiments as mentioned above with different feature missing ratios varying from 0.1 to 0.9. All of the three missing mechanisms are included. Figure \ref{fig:robustness} shows the imputation performance of METR-LA dataset regarding RMSE error. Our proposed GLPN shows consistent lowest imputation errors for different data missing levels. As the missing ratio increases, GLPN maintains a fairly stable performance in both MCAR and MAR mechanisms. Although the missingness of whole observations is more challenging (under MNAR), GLPN is still able to produce satisfactory imputation for unseen observations.

\subsubsection{Ablation Study}

In this subsection, we investigate the influence of two branches of GLPN on imputation performance, i.e. the residual network and the pyramid autoencoder. We test the variants of GLPN without residual network (denoted by\textit{ GLPN w/o R}) and U-shaped autoencoder (denoted by\textit{ GLPN w/o P}) separately. Under MCAR mechanisms, the corresponding results (RMSE) are shown in Table \ref{ablation}. We observe that the imputation error RMSE decreases on average 32\% without residual network. Meanwhile, the U-shaped autoencoder reduces the average RMSE by 35\%. By comparing the full GLPN model and two variants, it confirms that both residual network and U-shaped autoencoder branch can help improve the imputation performance.

\begin{table}[htb]
\centering
\small
\caption{Imputation Performance (MAE) on multi-graph datasets.}
\vspace{-3mm}
\resizebox{0.82\columnwidth}{!}{
\begin{tabular}{c|ccc}
\toprule
               & \textbf{SYNTHIE} & \textbf{PROTEINS} & \textbf{FRANKENSTEIN} \\ \midrule
\textbf{Mean}  & 2.239            & 17.925                 & 0.4496                \\

\textbf{Soft}  & 1.358            & 7.037                  & 0.0524                \\
\textbf{Mice}  & 1.297            & 3.482                  & 0.0513                \\
\textbf{OT}    & 2.027            & 9.907                  & 0.0632                \\
\textbf{GAIN}  & 1.664            & 7.003                  & 0.0523                \\
\midrule
\textbf{KNN}   & 2.152            & 6.758                  & 0.3144                \\
\textbf{GRAPE} & 1.579            & 6.236                  & 0.0588                \\
\textbf{GCN}   & 2.276            & 7.099                  & 0.0683                \\
\textbf{GDN}   & 1.543            & 6.755                  & 0.0509                \\
\textbf{GLPN}  & \textbf{1.209}    & \textbf{2.812}        & \textbf{0.0468}      \\\bottomrule
\end{tabular}
}
\label{table:non-sensor}
\vspace{-4mm}
\end{table}

\begin{table*}[tb]
\caption{Experiments with different draft strategies on continuous sensor dataset imputation. The table shows the RMSE normalized by the performance of mean imputation.}
\centering
\vspace{-3mm}
\resizebox{0.88\textwidth}{!}{
\begin{tabular}{c|ccc|ccc|ccc|ccc}
\toprule
 &
  \multicolumn{3}{c|}{\textbf{METR-LA}} &
  \multicolumn{3}{c|}{\textbf{NREL}} &
  \multicolumn{3}{c|}{\textbf{PEMS}} &
  \multicolumn{3}{c}{\textbf{SeData}} \\
 &
\textbf{  MCAR} &
\textbf{  MAR} &
 \textbf{ MNAR} &
  \textbf{MCAR} &
 \textbf{ MAR} &
 \textbf{ MNAR} &
\textbf{  MCAR} &
\textbf{  MAR} &
\textbf{  MNAR }&
\textbf{  MCAR} &
\textbf{  MAR} &
  \textbf{MNAR} \\ \midrule
\textbf{Mean} &
  1.000 &
  1.000 &
  1.000 &
  1.000 &
  1.000 &
  1.000 &
  1.000 &
  1.000 &
 1.000 &
  1.000 &
  1.000 &
  1.000 \\
\textbf{Mean+GLPN} &
  \textbf{0.343} &
  \textbf{0.298} &
  \textbf{0.401} &
  \textbf{0.127} &
  \textbf{0.095} &
  \textbf{0.198} &
  \textbf{0.224} &
  \textbf{0.182} &
  \textbf{0.569} &
  \textbf{0.428} &
  \textbf{0.589} &
  \textbf{0.589} \\ \midrule
\textbf{Soft} &
  0.389 &
  0.529 &
  0.440 &
  0.174 &
  0.360 &
  0.552 &
  0.617 &
  0.886 &
  0.630 &
  0.574 &
  0.900 &
  0.900 \\
\textbf{Soft+GLPN} &
  \textbf{0.233} &
  \textbf{0.204} &
  \textbf{0.584} &
  \textbf{0.076} &
  \textbf{0.085} &
  \textbf{0.223} &
  \textbf{0.188} &
  \textbf{0.188} &
  \textbf{0.695} &
  \textbf{0.324} &
  \textbf{0.670} &
  \textbf{0.670} \\ \midrule
\textbf{KNN} &
  0.805 &
  0.697 &
  0.521 &
  0.764 &
  0.667 &
  0.275 &
  0.905 &
  0.751 &
  0.758 &
  0.645 &
  0.617 &
  0.617 \\
\textbf{KNN+GLPN} &
  \textbf{0.349} &
  \textbf{0.305} &
  \textbf{0.382} &
  \textbf{0.117} &
  \textbf{0.086} &
  \textbf{0.224} &
  \textbf{0.225} &
  \textbf{0.184} &
  \textbf{0.676} &
  \textbf{0.334} &
  \textbf{0.595} &
  \textbf{0.595} \\ \midrule
\textbf{GCN} &
  0.595 &
  0.722 &
  0.450 &
  0.364 &
  0.616 &
  0.538 &
  0.642 &
  0.847 &
  0.728 &
  0.537 &
  0.735 &
  0.735 \\
\textbf{DGCN+GLPN} &
  \textbf{0.270} &
  \textbf{0.237} &
  \textbf{0.337} &
  \textbf{0.051} &
  \textbf{0.043} &
  \textbf{0.135} &
  \textbf{0.182} &
  \textbf{0.195} &
  \textbf{0.538} &
  \textbf{0.327} &
  \textbf{0.469} &
  \textbf{0.469} \\ \midrule
\end{tabular}}
\label{table:draft}
\end{table*}

\subsection{Evaluation on Heterophily Setting}

Real-word graphs do not always obey the homophily assumption that similar features or same class labels are linked together. For heterophily setting, linked nodes have dissimilar features and different class labels, which cause graph signal contains more energy at high-frequency components \cite{zhu2020beyond}. In order to test the ability to capture both low- and high-frequency components of our model, we conduct experiments for feature imputation and node classification on different homophily ratio graphs.

Given a graph $G = \{\mathcal{V}, \mathcal{E}\}$ and node label vector $y$, the node homophily ratio is defined as the average proportion of the neighbors with the same class of each node \cite{zheng2022graph}:
\begin{equation}
    h = \frac{1}{|\mathcal{V}|}\sum \limits_{v\in \mathcal{V}} \frac{|\{u \in \mathcal{N}:y_v=y_u\}|}{|\mathcal{N}(v)|}.
    \label{eq:homo_ratio}
\end{equation}

The homophily ratio $h$ of six single-graph datasets ranges from 0.06 to 0.81. We drop 80\% node features through MCAR mechanism, then impute the missing data and use 1-layer GCN for downstream classification. We report MAE and average accuracy (ACC) for imputation and classification, respectively. Table \ref{table:non-sensor1} gives detailed results. 

Overall, we observe that our model has the best performance in heterophily and homophily settings. GLPN reduces MAE on imputation tasks by an average of 26\% and improves accuracy on classification tasks by an average of 17\% over other models. We also observe that GDN achieves competitive performance in heterophily setting. This is likely due to GDN work as a high frequency amplifier which can capture the feature distribution under heterophily. 
However, in homophily, neighbors are likely to have similar features, so methods follow the homophily assumption (e.g., VGAE) has better performance than GDN.
Therefore, our method is able to maintain the same level of performance under heterophily and homophily settings through the design of the U-shaped autoencoder and special residual layers.

\subsection{Evaluation on Multi-graph Datasets}

We additionally test our model on several multi-graph datasets (i.e., SYNTHIE, PROTEINS, FRANKENSTEIN). In particular, the protein dataset, PROTEINS, is in mixed-data setting with both continuous features and discrete features.

We test our model under the MCAR mechanism with 20\% missing ratio and report the imputation results in Table \ref{table:non-sensor}. From these results, we can see that our proposed GLPN achieves the best imputation performance, which indicates that GLPN could be widely applied for different graph data imputation scenarios. Especially, for PROTEINS, GLPN is still able to beat other baseline methods in the mix-data setting.

\begin{figure}[tb]
  \centering

  \includegraphics[width=0.95\columnwidth]{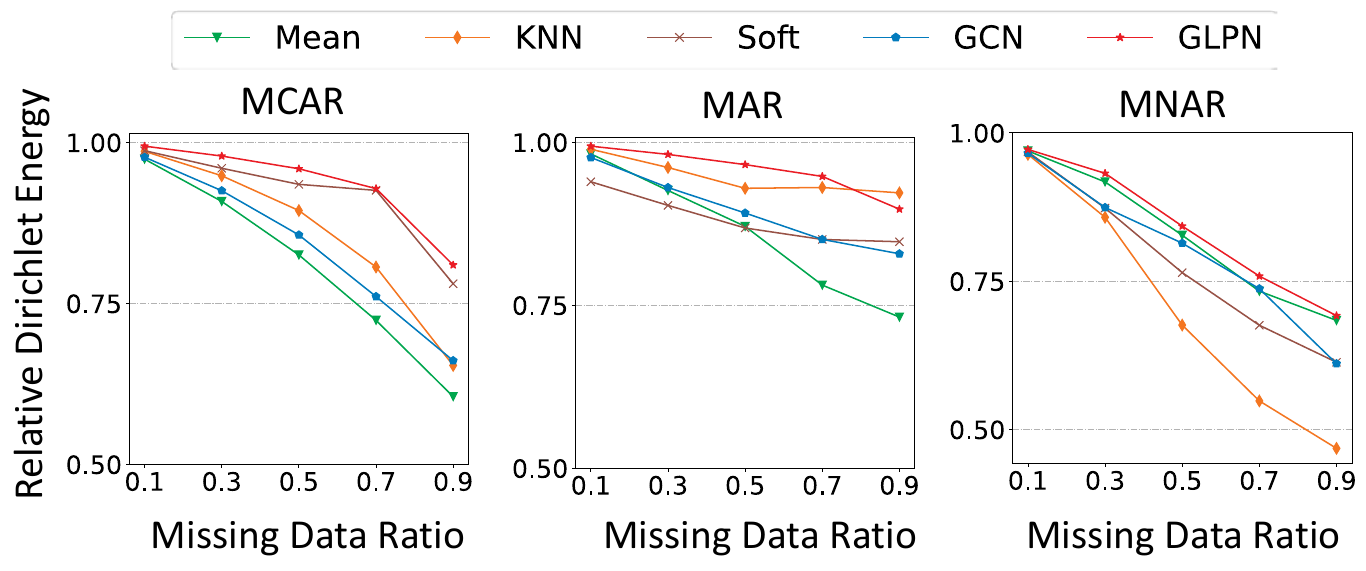}
\vspace{-5mm}
  \caption{Relative Dirichlet energy for METR-LA imputation with different data missing ratios. The Dirichlet energy is normalized by the ground truth features, whose relative energy equals 1.}

  \label{fig:energy_ratio}
  \vspace{-3mm}
\end{figure}

\subsection{Experiments with Different Draft Strategies}
Given that the draft imputation used to start the model matters a lot, we additionally consider four draft imputation methods, including two structure-free methods (MEAN, Soft) and two structure-based methods (K-hop nearest neighbors, GCN). Note that the model described in this paper utilizes a variant of Graph Convolutional Networks (i.e. DGCN) as the draft component. We report the normalized results relative to the performance of the MEAN imputation in Table \ref{table:draft}. Among these strategies, GCN based methods provide the best imputation performance on average. Given that the draft step is not the focus of this work, we only report the best model with DGCN drafter in the paper. It is worth noting that based on all these draft strategies, GLPN could significantly improve the final imputation performance via an energy-preservation refine step.

\subsection{Dirichlet Energy Maintenance}

To validate the Dirichlet energy maintenance capability of GLPN, we compare our model with several baselines in Figure \ref{fig:energy_ratio}. 
The experiments are conducted on METR-LA dataset with different data missing ratios. 
The Dirichlet energy is normalized by the ground truth feature, which is the target of imputation. 
Among all the baselines, we can observe that the imputation energy of GLPN decreases the slowest with an increasing missing ratio for all three types of missing mechanisms, which also reveals the robustness of our imputation model. 


To illustrate the correlation between Dirichlet energy and imputation performance, we also report the imputation performance and the relative Dirichlet energy gap between the model prediction $\bm{\hat{X}}$ and the ground truth $\bm{X}$, i.e., $\Delta E = \frac{E_D(\bm{\hat{X}})-E_D(\bm{X})}{E_D(\bm{X})}$ in Table \ref{table:energy_corr}. 
Among all these methods, our proposed GLPN achieves the best imputation performance as well as the least Dirichlet energy gap, consistent with the motivation of this work.

\begin{table}[tb]
\centering

  \caption{Correlations between the imputation error and the Dirichlet energy gap on METR-LA dataset. $\Delta E$ denotes the relative Dirichlet energy gap.}
\vspace{-3mm}
\resizebox{0.75\columnwidth}{!}{
\begin{tabular}{cc|cccc}
\toprule
\multicolumn{2}{c|}{\textbf{Missing Ratio}}     & \textbf{10\%}   & \textbf{30\%}   & \textbf{50\%  }  & \textbf{70\% }   \\ \midrule
\multirow{2}{*}{\textbf{Mean}} &    $\Delta E$           & -1.8\% & -7.4\% & -13.0\% & -22.0\% \\
                               & RMSE & 14.13  & 13.74  & 14.94   & 16.52   \\ \midrule
\multirow{2}{*}{\textbf{KNN}}  &       $\Delta E$         & -1.1\% & -3.9\% & -7.1\%  & -7.2\%  \\
                               & {RMSE} & 11.29  & 11.89  & 11.98   & 14.13   \\ \midrule
\multirow{2}{*}{\textbf{GCN}}           &      $\Delta E$          & -2.4\% & -6.9\% & -10.9\% & -14.9\% \\
                               & {RMSE} & 7.71   & 7.65   & 8.37    & 8.44    \\ \midrule
\multirow{2}{*}{\textbf{GLPN}}          &       $\Delta E$         & -0.7\% & -1.9\% & -3.5\%  & -5.3\%  \\
                               & {RMSE} & 4.70   & 4.58   & 4.94    & 5.49   \\ \bottomrule
\end{tabular}}

\label{table:energy_corr}
\vspace{-3mm}
\end{table}

\vspace{-2mm}
\section{Conclusion}\label{sec:conclusion}
In this paper, we present Graph Laplacian Pyramid Network (GLPN) for missing data imputation. We initialize the first study to discuss the ``draft-then-refine" imputation paradigm from the perspective of Dirichlet energy. Based on the ``draft-then-refine'' procedure, we develop a U-shaped autoencoder and residual network to refine the node representations based on draft imputation. We find that Dirichlet energy can be a principle to guide the design of the imputation model. We also theoretically demonstrate that our model has better energy maintenance ability. The experiments show that our model has significant improvements on several imputation tasks compared against state-of-the-art imputation approaches.

\begin{acks}
This work is funded by National Natural Science Foundation of China Grant No. 72371271, the Guangzhou Industrial Information and Intelligent Key Laboratory Project (No. 2024A03J0628), the Nansha Key Area Science and Technology Project (No. 2023ZD003), and Project No. 2021JC02X191.
\end{acks}
\appendix

\section{Proof for Equation (2)} 
\begin{proof}
Let $E_D(\bm X)=tr\left(\bm X^T \tilde{\Delta} \bm X\right)$ be the Dirichlet energy, $\bm X_a$ and $\bm X_b$ be two bounded feature matrices, and $B = \max\left(\parallel \bm X_a\parallel, \parallel\bm X_b\parallel\right)$ be the $\ell_2$ boundary. Consider that $E_D(\bm X)$ is a local Lipschitz function when $\bm X$ is bounded, we have:
\begin{equation*}
    \begin{aligned}
    \frac{|E_D(\bm X_a) - E_D(\bm X_b)|}{\parallel \bm X_a - \bm X_b \parallel} \leq & \sup_{\parallel \bm X\parallel \leq B} \parallel \frac{\partial E_D(\bm X)}{\partial \bm X} \parallel\\
    = & \sup_{\parallel \bm X\parallel \leq B} \parallel 2\tilde{\Delta} \bm X \parallel  \\  \leq &  ~ 2 \lambda_{max} B.
    \end{aligned}
\end{equation*}
\end{proof}

\vspace{-4mm}

\section{Proof of Proposition 3.2} 
\label{proof:proposition3}

\begin{proof}
We first consider the case that $\bm X$ only has one feature column, i.e., $\bm X \in \mathbb R^{N\times 1}$. According to Equation (1), if $\bm X_i$ and $\bm X_j$ are connected, the related term in the graph Dirichlet energy is
\begin{equation*}
\resizebox{0.55\columnwidth}{!}{$
    e_D\left(\bm X_i,\bm X_j\right)=\left(\dfrac{\bm X_{i}}{\sqrt{1+\bm D_{i,i}}}-\dfrac{\bm X_{j}}{\sqrt{1+\bm D_{j,j}}}\right)^2. $}
\end{equation*}

\noindent Using the first and second moment constraints of $\bm X_i$ and $\bm X_j$, the expectation of $e_D(\bm X_i,\bm X_j)$ can be calculated by
\begin{equation*}
\resizebox{\columnwidth}{!}{ $
    \mathbb E\left(e_D(\bm X_i,\bm X_j)\right)=\text{Var}\left(\dfrac{\bm X_{i}}{\sqrt{1+\bm D_{i,i}}}\right)+\text{Var}\left(\dfrac{\bm X_{j}}{\sqrt{1+\bm D_{j,j}}}\right) =
    \frac{1}{1+\bm D_{i,i}}+\frac{1}{1+\bm D_{j,j}}.$}
\end{equation*}

\noindent If $\bm X_i$ is missing and imputed by $\hat{\bm X}_{i}=\sum_{k\in S_k}\alpha_{k}\bm X_{k}$ while $\hat{\bm X_j}=\bm X_j$ is unchanged, we have
\begin{equation*}
\resizebox{0.82\columnwidth}{!}{$\begin{aligned}
\mathbb E\left(e_D(\hat{\bm X}_{i},\hat{\bm X_j})\right)\leq& \text{Var}\left(\dfrac{\sum_{k\in S_k}\alpha_{k}\bm X_{k}}{\sqrt{1+\bm D_{i,i}}}\right)+\text{Var}\left(\dfrac{\bm X_{j}}{\sqrt{1+\bm D_{j,j}}}\right)\\
\leq& \frac{\sum_{k\in S_k}\alpha_{k}^2}{1+\bm D_{i,i}}+\frac{1}{1+\bm D_{j,j}} \leq \frac{1}{1+\bm D_{i,i}}+\frac{1}{1+\bm D_{j,j}}.
\end{aligned}$}
\end{equation*}

\noindent On the other side, if both $\bm X_i$ and $\bm X_j$ are missing and imputed by $\hat{\bm X}_{i}=\sum_{k\in S_k}\alpha_{k}\bm X_{k}$ and $\hat{\bm X}_{j}=\sum_{k\in S_k}\beta_{k}\bm X_{k}$ respectively, we have

\begin{equation*}
\resizebox{0.85\columnwidth}{!}{$\begin{aligned}
\mathbb E\left(e_D(\hat{\bm X}_{i},\hat{\bm X_j})\right)\leq& \text{Var}\left(\dfrac{\sum_{k\in S_k}\alpha_{k}\bm X_{k}}{\sqrt{1+\bm D_{i,i}}}\right)+\text{Var}\left(\dfrac{\sum_{k\in S_k}\beta_{k}\bm X_{k}}{\sqrt{1+\bm D_{j,j}}}\right)\\
\leq& \frac{\sum_{k\in S_k}\alpha_{k}^2}{1+\bm D_{i,i}}+\frac{\sum_{k\in S_k}\beta_{k}^2}{1+\bm D_{j,j}} 
\leq \frac{1}{1+\bm D_{i,i}}+\frac{1}{1+\bm D_{j,j}}. 
    \end{aligned}$}
\end{equation*}

\noindent In summary, we have
\begin{equation*}
\resizebox{\columnwidth}{!}{ $
    \mathbb E[E_D(\hat{\bm X})]=\sum_{\{(i,j)|\bm A_{ij}=1\}}E(e_D(\hat{\bm X}_{i},\hat{\bm X_j})) \leq \sum_{\{(i,j)|\bm A_{ij}=1\}}E(e_D(\bm X_{i},\bm X_j))\leq \mathbb E[E_D(\bm X)]. $}
\end{equation*}

\noindent Similarly, if $\bm X$ has more than one feature column, i.e., $\bm X \in \mathbb R^{N\times d}$, we have $\mathbb E[E_D(\hat{\bm X}_{i,:})]\leq \mathbb E[E_D(\bm X_{i,:})]$ for $0\leq i < d$, and thus $\mathbb E[E_D(\hat{\bm X})]\leq \mathbb E[E_D(\bm X)]$.







\end{proof}

\section{Proof for Proposition 5.1}  \label{sec:append2}



\begin{proof}
As illustrated in Section 5.2, for the proposed GLPN with one-level pooling-unpooling, the output of our model could be formulated by: 
\begin{equation*}
    \hat{\bm X}=\bm P_l \bm X_d+ \alpha \bm S\bm S^T\bm X_d.
\end{equation*}
We define $\bm Q \triangleq \bm P_l+ \alpha \bm S\bm S^T$, and then simplify the above expression as 
\begin{equation*}
    \hat{\bm X}=\bm Q \bm X_d  .
\end{equation*}
We represent the decomposition of matrix $\bm Q$ as: $\bm Q = \bm U \bm \Lambda \bm U^{T}$, where the columns of $\bm U$ constitute
an orthonormal basis of eigenvectors of $\bm Q$, and the diagonal matrix $\bm \Lambda$ is comprised of the corresponding eigenvalues of $\bm Q$. 

\noindent For $E_D(\hat{\bm X})$, we have:
\begin{equation*}
\begin{split}
E_D(\hat{\bm X})& =\operatorname{tr} \left(( \bm U \bm \Lambda\bm U^T \bm X_d)^T \tilde{\Delta} (\bm U \bm \Lambda\bm U^T \bm X_d) \right) \\
& = \operatorname{tr}\left(\bm U^T \bm X_d \bm X^T_d \bm U \bm \Lambda\bm U^T \tilde{\Delta} \bm U \bm \Lambda\right) \\
& \geq \delta \operatorname{tr}\left(\bm U^T \bm X_d \bm X^T_d \bm U \bm \Lambda\bm U^T \tilde{\Delta} \bm U\right) \\
& \geq {\delta}^2 \operatorname{tr}\left(\bm X_d^T \tilde{\Delta} \bm X_d\right) \\
& = {\delta}^2 E_D(\bm X_d),
\end{split}
\end{equation*}
where $\delta$ is the minimum eigenvalue of matrix $\bm Q$. 
Recalling $\bm Q\triangleq \bm P_l+ \alpha \bm S\bm S^T$ and $\bm P_l\triangleq\bm I+\tilde{\Delta}$, then we have $\delta = 1+C_{min}$, where $C_{min}$ is the minimum eigenvalue of matrix $\tilde{\Delta} + \alpha \bm S\bm S^T$.

\noindent In summary, we have
\begin{equation*}
 (1 + C_{min})^2 E_D(\bm X_d) \leq E_D (\hat{\bm X})   ,
\end{equation*}
where $C_{min}$ is the minimum eigenvalue of matrix $\tilde{\Delta}+\alpha \bm S\bm S^T$.

\end{proof}

\section{Proposition 5.1 with Higher-order Maclaurin Series Approximation }
Although we adopt the first-order Maclaurin series approximation of GDN in analysis, we can obtain similar results in the higher-order approximation. For example, using the second-order Maclaurin Series approximation, Equation (11) would be $$\hat{\bm X}=\tilde{\bm P}_l  \bm X_d+ \alpha  \bm S \bm S^T \bm X_d,$$ where $ \tilde{\bm P}_l= \bm I+\tilde{\Delta}+\frac{1}{2}{\tilde{\Delta}}^2$. Then, we have a similar conclusion for Proposition 5.1:
$$(1 + \tilde{C}_{min})^2 E_D(\bm X_d) \leq E_D (\hat{\bm X}), $$ 
where $\tilde{C}_{min}$ is the minimum eigenvalue of matrix $\tilde{\Delta}+\frac{1}{2}{\tilde{\Delta}}^2+\alpha \bm S \bm S^T$.

\bibliographystyle{ACM-Reference-Format}
\bibliography{ref}


\begin{thebibliography}{66}


\ifx \showCODEN    \undefined \def \showCODEN     #1{\unskip}     \fi
\ifx \showDOI      \undefined \def \showDOI       #1{#1}\fi
\ifx \showISBNx    \undefined \def \showISBNx     #1{\unskip}     \fi
\ifx \showISBNxiii \undefined \def \showISBNxiii  #1{\unskip}     \fi
\ifx \showISSN     \undefined \def \showISSN      #1{\unskip}     \fi
\ifx \showLCCN     \undefined \def \showLCCN      #1{\unskip}     \fi
\ifx \shownote     \undefined \def \shownote      #1{#1}          \fi
\ifx \showarticletitle \undefined \def \showarticletitle #1{#1}   \fi
\ifx \showURL      \undefined \def \showURL       {\relax}        \fi
\providecommand\bibfield[2]{#2}
\providecommand\bibinfo[2]{#2}
\providecommand\natexlab[1]{#1}
\providecommand\showeprint[2][]{arXiv:#2}

\bibitem[Anwar and Barnes(2020)]%
        {anwar2020densely}
\bibfield{author}{\bibinfo{person}{Saeed Anwar} {and} \bibinfo{person}{Nick Barnes}.} \bibinfo{year}{2020}\natexlab{}.
\newblock \showarticletitle{Densely residual laplacian super-resolution}.
\newblock \bibinfo{journal}{\emph{IEEE Transactions on Pattern Analysis and Machine Intelligence}} \bibinfo{volume}{44}, \bibinfo{number}{3} (\bibinfo{year}{2020}), \bibinfo{pages}{1192--1204}.
\newblock


\bibitem[Audigier et~al\mbox{.}(2016)]%
        {audigier2016multiple}
\bibfield{author}{\bibinfo{person}{Vincent Audigier}, \bibinfo{person}{Fran{\c{c}}ois Husson}, {and} \bibinfo{person}{Julie Josse}.} \bibinfo{year}{2016}\natexlab{}.
\newblock \showarticletitle{Multiple imputation for continuous variables using a Bayesian principal component analysis}.
\newblock \bibinfo{journal}{\emph{Journal of statistical computation and simulation}} \bibinfo{volume}{86}, \bibinfo{number}{11} (\bibinfo{year}{2016}), \bibinfo{pages}{2140--2156}.
\newblock


\bibitem[Bechny et~al\mbox{.}(2021)]%
        {bechny2021missing}
\bibfield{author}{\bibinfo{person}{Michal Bechny}, \bibinfo{person}{Florian Sobieczky}, \bibinfo{person}{J{\"u}rgen Zeindl}, {and} \bibinfo{person}{Lisa Ehrlinger}.} \bibinfo{year}{2021}\natexlab{}.
\newblock \showarticletitle{Missing Data Patterns: From Theory to an Application in the Steel Industry}. In \bibinfo{booktitle}{\emph{33rd International Conference on Scientific and Statistical Database Management}}. \bibinfo{pages}{214--219}.
\newblock


\bibitem[Bertsimas et~al\mbox{.}(2017)]%
        {bertsimas2017predictive}
\bibfield{author}{\bibinfo{person}{Dimitris Bertsimas}, \bibinfo{person}{Colin Pawlowski}, {and} \bibinfo{person}{Ying~Daisy Zhuo}.} \bibinfo{year}{2017}\natexlab{}.
\newblock \showarticletitle{From predictive methods to missing data imputation: an optimization approach}.
\newblock \bibinfo{journal}{\emph{The Journal of Machine Learning Research}} \bibinfo{volume}{18}, \bibinfo{number}{1} (\bibinfo{year}{2017}), \bibinfo{pages}{7133--7171}.
\newblock


\bibitem[Borgwardt et~al\mbox{.}(2005)]%
        {borgwardt2005protein}
\bibfield{author}{\bibinfo{person}{Karsten~M Borgwardt}, \bibinfo{person}{Cheng~Soon Ong}, \bibinfo{person}{Stefan Sch{\"o}nauer}, \bibinfo{person}{SVN Vishwanathan}, \bibinfo{person}{Alex~J Smola}, {and} \bibinfo{person}{Hans-Peter Kriegel}.} \bibinfo{year}{2005}\natexlab{}.
\newblock \showarticletitle{Protein function prediction via graph kernels}.
\newblock \bibinfo{journal}{\emph{Bioinformatics}} \bibinfo{volume}{21}, \bibinfo{number}{suppl\_1} (\bibinfo{year}{2005}), \bibinfo{pages}{i47--i56}.
\newblock


\bibitem[Bose et~al\mbox{.}(2013a)]%
        {bose2013modified}
\bibfield{author}{\bibinfo{person}{Shilpi Bose}, \bibinfo{person}{Chandra Das}, \bibinfo{person}{Tamaghna Gangopadhyay}, {and} \bibinfo{person}{Samiran Chattopadhyay}.} \bibinfo{year}{2013}\natexlab{a}.
\newblock \showarticletitle{A modified local least squares-based missing value estimation method in microarray gene expression data}. In \bibinfo{booktitle}{\emph{2013 2nd International Conference on Advanced Computing, Networking and Security}}. IEEE, \bibinfo{pages}{18--23}.
\newblock


\bibitem[Bose et~al\mbox{.}(2013b)]%
        {BoseDGC13}
\bibfield{author}{\bibinfo{person}{Shilpi Bose}, \bibinfo{person}{Chandra Das}, \bibinfo{person}{Tamaghna Gangopadhyay}, {and} \bibinfo{person}{Samiran Chattopadhyay}.} \bibinfo{year}{2013}\natexlab{b}.
\newblock \showarticletitle{A Modified Local Least Squares-Based Missing Value Estimation Method in Microarray Gene Expression Data}. In \bibinfo{booktitle}{\emph{2013 2nd International Conference on Advanced Computing, Networking and Security, Mangalore, India, December 15-17, 2013}}. \bibinfo{publisher}{{IEEE}}, \bibinfo{pages}{18--23}.
\newblock


\bibitem[Burt and Adelson(1987)]%
        {burt1987laplacian}
\bibfield{author}{\bibinfo{person}{Peter~J Burt} {and} \bibinfo{person}{Edward~H Adelson}.} \bibinfo{year}{1987}\natexlab{}.
\newblock \showarticletitle{The Laplacian pyramid as a compact image code}.
\newblock In \bibinfo{booktitle}{\emph{Readings in computer vision}}. \bibinfo{publisher}{Elsevier}, \bibinfo{pages}{671--679}.
\newblock


\bibitem[Cai and Wang(2020)]%
        {cai2020note}
\bibfield{author}{\bibinfo{person}{Chen Cai} {and} \bibinfo{person}{Yusu Wang}.} \bibinfo{year}{2020}\natexlab{}.
\newblock \showarticletitle{A note on over-smoothing for graph neural networks}.
\newblock \bibinfo{journal}{\emph{arXiv preprint arXiv:2006.13318}} (\bibinfo{year}{2020}).
\newblock


\bibitem[Cai et~al\mbox{.}(2010)]%
        {cai2010singular}
\bibfield{author}{\bibinfo{person}{Jian-Feng Cai}, \bibinfo{person}{Emmanuel~J Cand{\`e}s}, {and} \bibinfo{person}{Zuowei Shen}.} \bibinfo{year}{2010}\natexlab{}.
\newblock \showarticletitle{A singular value thresholding algorithm for matrix completion}.
\newblock \bibinfo{journal}{\emph{SIAM Journal on optimization}} \bibinfo{volume}{20}, \bibinfo{number}{4} (\bibinfo{year}{2010}), \bibinfo{pages}{1956--1982}.
\newblock


\bibitem[Cui et~al\mbox{.}(2019)]%
        {cui2019traffic}
\bibfield{author}{\bibinfo{person}{Zhiyong Cui}, \bibinfo{person}{Kristian Henrickson}, \bibinfo{person}{Ruimin Ke}, {and} \bibinfo{person}{Yinhai Wang}.} \bibinfo{year}{2019}\natexlab{}.
\newblock \showarticletitle{Traffic graph convolutional recurrent neural network: A deep learning framework for network-scale traffic learning and forecasting}.
\newblock \bibinfo{journal}{\emph{IEEE Transactions on Intelligent Transportation Systems}} (\bibinfo{year}{2019}).
\newblock


\bibitem[Cui et~al\mbox{.}(2018)]%
        {cui2018deep}
\bibfield{author}{\bibinfo{person}{Zhiyong Cui}, \bibinfo{person}{Ruimin Ke}, {and} \bibinfo{person}{Yinhai Wang}.} \bibinfo{year}{2018}\natexlab{}.
\newblock \showarticletitle{Deep Bidirectional and Unidirectional LSTM Recurrent Neural Network for Network-wide Traffic Speed Prediction}.
\newblock \bibinfo{journal}{\emph{arXiv preprint arXiv:1801.02143}} (\bibinfo{year}{2018}).
\newblock


\bibitem[De~Silva and Perera(2016)]%
        {de2016missing}
\bibfield{author}{\bibinfo{person}{Hiroshi De~Silva} {and} \bibinfo{person}{A~Shehan Perera}.} \bibinfo{year}{2016}\natexlab{}.
\newblock \showarticletitle{Missing data imputation using Evolutionary k-Nearest neighbor algorithm for gene expression data}. In \bibinfo{booktitle}{\emph{2016 Sixteenth International Conference on Advances in ICT for Emerging Regions (ICTer)}}. IEEE, \bibinfo{pages}{141--146}.
\newblock


\bibitem[Dempster et~al\mbox{.}(1977)]%
        {dempster1977maximum}
\bibfield{author}{\bibinfo{person}{Arthur~P Dempster}, \bibinfo{person}{Nan~M Laird}, {and} \bibinfo{person}{Donald~B Rubin}.} \bibinfo{year}{1977}\natexlab{}.
\newblock \showarticletitle{Maximum likelihood from incomplete data via the EM algorithm}.
\newblock \bibinfo{journal}{\emph{Journal of the Royal Statistical Society: Series B (Methodological)}} \bibinfo{volume}{39}, \bibinfo{number}{1} (\bibinfo{year}{1977}), \bibinfo{pages}{1--22}.
\newblock


\bibitem[Emmanuel et~al\mbox{.}(2021)]%
        {EmmanuelMMSMT21}
\bibfield{author}{\bibinfo{person}{Tlamelo Emmanuel}, \bibinfo{person}{Thabiso~M. Maupong}, \bibinfo{person}{Dimane Mpoeleng}, \bibinfo{person}{Thabo Semong}, \bibinfo{person}{Banyatsang Mphago}, {and} \bibinfo{person}{Oteng Tabona}.} \bibinfo{year}{2021}\natexlab{}.
\newblock \showarticletitle{A survey on missing data in machine learning}.
\newblock \bibinfo{journal}{\emph{J. Big Data}} \bibinfo{volume}{8}, \bibinfo{number}{1} (\bibinfo{year}{2021}), \bibinfo{pages}{140}.
\newblock
\urldef\tempurl%
\url{https://doi.org/10.1186/s40537-021-00516-9}
\showDOI{\tempurl}


\bibitem[Gao and Ji(2019)]%
        {gao2019graph}
\bibfield{author}{\bibinfo{person}{Hongyang Gao} {and} \bibinfo{person}{Shuiwang Ji}.} \bibinfo{year}{2019}\natexlab{}.
\newblock \showarticletitle{Graph u-nets}. In \bibinfo{booktitle}{\emph{international conference on machine learning}}. PMLR, \bibinfo{pages}{2083--2092}.
\newblock


\bibitem[Gao et~al\mbox{.}(2023)]%
        {gao2023handling}
\bibfield{author}{\bibinfo{person}{Ziqi Gao}, \bibinfo{person}{Yifan Niu}, \bibinfo{person}{Jiashun Cheng}, \bibinfo{person}{Jianheng Tang}, \bibinfo{person}{Lanqing Li}, \bibinfo{person}{Tingyang Xu}, \bibinfo{person}{Peilin Zhao}, \bibinfo{person}{Fugee Tsung}, {and} \bibinfo{person}{Jia Li}.} \bibinfo{year}{2023}\natexlab{}.
\newblock \showarticletitle{Handling missing data via max-entropy regularized graph autoencoder}. In \bibinfo{booktitle}{\emph{Proceedings of the AAAI Conference on Artificial Intelligence}}, Vol.~\bibinfo{volume}{37}. \bibinfo{pages}{7651--7659}.
\newblock


\bibitem[Garc{\'\i}a-Laencina et~al\mbox{.}(2010)]%
        {garcia2010pattern}
\bibfield{author}{\bibinfo{person}{Pedro~J Garc{\'\i}a-Laencina}, \bibinfo{person}{Jos{\'e}-Luis Sancho-G{\'o}mez}, {and} \bibinfo{person}{An{\'\i}bal~R Figueiras-Vidal}.} \bibinfo{year}{2010}\natexlab{}.
\newblock \showarticletitle{Pattern classification with missing data: a review}.
\newblock \bibinfo{journal}{\emph{Neural Computing and Applications}} \bibinfo{volume}{19}, \bibinfo{number}{2} (\bibinfo{year}{2010}), \bibinfo{pages}{263--282}.
\newblock


\bibitem[Garc{\'\i}a-Plaza et~al\mbox{.}(2016)]%
        {garcia2016using}
\bibfield{author}{\bibinfo{person}{Alberto~P Garc{\'\i}a-Plaza}, \bibinfo{person}{V{\'\i}ctor Fresno}, \bibinfo{person}{Raquel~Mart{\'\i}nez Unanue}, {and} \bibinfo{person}{Arkaitz Zubiaga}.} \bibinfo{year}{2016}\natexlab{}.
\newblock \showarticletitle{Using fuzzy logic to leverage HTML markup for web page representation}.
\newblock \bibinfo{journal}{\emph{IEEE Transactions on Fuzzy Systems}} \bibinfo{volume}{25}, \bibinfo{number}{4} (\bibinfo{year}{2016}), \bibinfo{pages}{919--933}.
\newblock


\bibitem[Gondara and Wang(2018)]%
        {gondara2017multiple}
\bibfield{author}{\bibinfo{person}{Lovedeep Gondara} {and} \bibinfo{person}{Ke Wang}.} \bibinfo{year}{2018}\natexlab{}.
\newblock \showarticletitle{Mida: Multiple imputation using denoising autoencoders}. In \bibinfo{booktitle}{\emph{Pacific-Asia conference on knowledge discovery and data mining}}. Springer, \bibinfo{pages}{260--272}.
\newblock


\bibitem[Hastie et~al\mbox{.}(2015)]%
        {hastie2015matrix}
\bibfield{author}{\bibinfo{person}{Trevor Hastie}, \bibinfo{person}{Rahul Mazumder}, \bibinfo{person}{Jason~D Lee}, {and} \bibinfo{person}{Reza Zadeh}.} \bibinfo{year}{2015}\natexlab{}.
\newblock \showarticletitle{Matrix completion and low-rank SVD via fast alternating least squares}.
\newblock \bibinfo{journal}{\emph{The Journal of Machine Learning Research}} \bibinfo{volume}{16}, \bibinfo{number}{1} (\bibinfo{year}{2015}), \bibinfo{pages}{3367--3402}.
\newblock


\bibitem[Huang et~al\mbox{.}(2020)]%
        {huang2020tackling}
\bibfield{author}{\bibinfo{person}{Wenbing Huang}, \bibinfo{person}{Yu Rong}, \bibinfo{person}{Tingyang Xu}, \bibinfo{person}{Fuchun Sun}, {and} \bibinfo{person}{Junzhou Huang}.} \bibinfo{year}{2020}\natexlab{}.
\newblock \showarticletitle{Tackling over-smoothing for general graph convolutional networks}.
\newblock \bibinfo{journal}{\emph{arXiv preprint arXiv:2008.09864}} (\bibinfo{year}{2020}).
\newblock


\bibitem[Ivanov et~al\mbox{.}(2019)]%
        {ivanov2019variational}
\bibfield{author}{\bibinfo{person}{O Ivanov}, \bibinfo{person}{M Figurnov}, {and} \bibinfo{person}{D Vetrov}.} \bibinfo{year}{2019}\natexlab{}.
\newblock \showarticletitle{Variational autoencoder with arbitrary conditioning}. In \bibinfo{booktitle}{\emph{7th International Conference on Learning Representations, ICLR 2019}}.
\newblock


\bibitem[Keerin and Boongoen(2021)]%
        {keerin2021improved}
\bibfield{author}{\bibinfo{person}{Phimmarin Keerin} {and} \bibinfo{person}{Tossapon Boongoen}.} \bibinfo{year}{2021}\natexlab{}.
\newblock \showarticletitle{Improved knn imputation for missing values in gene expression data}.
\newblock \bibinfo{journal}{\emph{Computers, Materials and Continua}} \bibinfo{volume}{70}, \bibinfo{number}{2} (\bibinfo{year}{2021}), \bibinfo{pages}{4009--4025}.
\newblock


\bibitem[Kim et~al\mbox{.}(2004)]%
        {kim2004reuse}
\bibfield{author}{\bibinfo{person}{Ki-Yeol Kim}, \bibinfo{person}{Byoung-Jin Kim}, {and} \bibinfo{person}{Gwan-Su Yi}.} \bibinfo{year}{2004}\natexlab{}.
\newblock \showarticletitle{Reuse of imputed data in microarray analysis increases imputation efficiency}.
\newblock \bibinfo{journal}{\emph{BMC bioinformatics}} \bibinfo{volume}{5}, \bibinfo{number}{1} (\bibinfo{year}{2004}), \bibinfo{pages}{1--9}.
\newblock


\bibitem[Kipf and Welling(2016)]%
        {kipf2016variational}
\bibfield{author}{\bibinfo{person}{Thomas~N Kipf} {and} \bibinfo{person}{Max Welling}.} \bibinfo{year}{2016}\natexlab{}.
\newblock \showarticletitle{Variational graph auto-encoders}.
\newblock \bibinfo{journal}{\emph{arXiv preprint arXiv:1611.07308}} (\bibinfo{year}{2016}).
\newblock


\bibitem[Kipf and Welling(2017)]%
        {kipf2016semi}
\bibfield{author}{\bibinfo{person}{Thomas~N Kipf} {and} \bibinfo{person}{Max Welling}.} \bibinfo{year}{2017}\natexlab{}.
\newblock \showarticletitle{Semi-Supervised Classification with Graph Convolutional Networks}. In \bibinfo{booktitle}{\emph{International Conference on Learning Representations}}.
\newblock


\bibitem[Kyono et~al\mbox{.}(2021)]%
        {kyono2021miracle}
\bibfield{author}{\bibinfo{person}{Trent Kyono}, \bibinfo{person}{Yao Zhang}, \bibinfo{person}{Alexis Bellot}, {and} \bibinfo{person}{Mihaela van~der Schaar}.} \bibinfo{year}{2021}\natexlab{}.
\newblock \showarticletitle{MIRACLE: Causally-Aware Imputation via Learning Missing Data Mechanisms}.
\newblock \bibinfo{journal}{\emph{Advances in Neural Information Processing Systems}}  \bibinfo{volume}{34} (\bibinfo{year}{2021}).
\newblock


\bibitem[Lai et~al\mbox{.}(2017)]%
        {lai2017deep}
\bibfield{author}{\bibinfo{person}{Wei-Sheng Lai}, \bibinfo{person}{Jia-Bin Huang}, \bibinfo{person}{Narendra Ahuja}, {and} \bibinfo{person}{Ming-Hsuan Yang}.} \bibinfo{year}{2017}\natexlab{}.
\newblock \showarticletitle{Deep laplacian pyramid networks for fast and accurate super-resolution}. In \bibinfo{booktitle}{\emph{Proceedings of the IEEE conference on computer vision and pattern recognition}}. \bibinfo{pages}{624--632}.
\newblock


\bibitem[Lai et~al\mbox{.}(2018)]%
        {lai2018fast}
\bibfield{author}{\bibinfo{person}{Wei-Sheng Lai}, \bibinfo{person}{Jia-Bin Huang}, \bibinfo{person}{Narendra Ahuja}, {and} \bibinfo{person}{Ming-Hsuan Yang}.} \bibinfo{year}{2018}\natexlab{}.
\newblock \showarticletitle{Fast and accurate image super-resolution with deep laplacian pyramid networks}.
\newblock \bibinfo{journal}{\emph{IEEE transactions on pattern analysis and machine intelligence}} \bibinfo{volume}{41}, \bibinfo{number}{11} (\bibinfo{year}{2018}), \bibinfo{pages}{2599--2613}.
\newblock


\bibitem[Lakshminarayan et~al\mbox{.}(1999)]%
        {lakshminarayan1999imputation}
\bibfield{author}{\bibinfo{person}{Kamakshi Lakshminarayan}, \bibinfo{person}{Steven~A Harp}, {and} \bibinfo{person}{Tariq Samad}.} \bibinfo{year}{1999}\natexlab{}.
\newblock \showarticletitle{Imputation of missing data in industrial databases}.
\newblock \bibinfo{journal}{\emph{Applied intelligence}} \bibinfo{volume}{11}, \bibinfo{number}{3} (\bibinfo{year}{1999}), \bibinfo{pages}{259--275}.
\newblock


\bibitem[Le~Morvan et~al\mbox{.}(2021)]%
        {le2021sa}
\bibfield{author}{\bibinfo{person}{Marine Le~Morvan}, \bibinfo{person}{Julie Josse}, \bibinfo{person}{Erwan Scornet}, {and} \bibinfo{person}{Ga{\"e}l Varoquaux}.} \bibinfo{year}{2021}\natexlab{}.
\newblock \showarticletitle{What’sa good imputation to predict with missing values?}
\newblock \bibinfo{journal}{\emph{Advances in Neural Information Processing Systems}}  \bibinfo{volume}{34} (\bibinfo{year}{2021}).
\newblock


\bibitem[Lee et~al\mbox{.}(2019)]%
        {lee2019self}
\bibfield{author}{\bibinfo{person}{Junhyun Lee}, \bibinfo{person}{Inyeop Lee}, {and} \bibinfo{person}{Jaewoo Kang}.} \bibinfo{year}{2019}\natexlab{}.
\newblock \showarticletitle{Self-attention graph pooling}. In \bibinfo{booktitle}{\emph{International conference on machine learning}}. PMLR, \bibinfo{pages}{3734--3743}.
\newblock


\bibitem[Li et~al\mbox{.}(2021)]%
        {li2021deconvolutional}
\bibfield{author}{\bibinfo{person}{Jia Li}, \bibinfo{person}{Jiajin Li}, \bibinfo{person}{Yang Liu}, \bibinfo{person}{Jianwei Yu}, \bibinfo{person}{Yueting Li}, {and} \bibinfo{person}{Hong Cheng}.} \bibinfo{year}{2021}\natexlab{}.
\newblock \showarticletitle{Deconvolutional Networks on Graph Data}.
\newblock \bibinfo{journal}{\emph{Advances in Neural Information Processing Systems}}  \bibinfo{volume}{34} (\bibinfo{year}{2021}).
\newblock


\bibitem[Li et~al\mbox{.}(2019)]%
        {li2019semi}
\bibfield{author}{\bibinfo{person}{Jia Li}, \bibinfo{person}{Yu Rong}, \bibinfo{person}{Hong Cheng}, \bibinfo{person}{Helen Meng}, \bibinfo{person}{Wenbing Huang}, {and} \bibinfo{person}{Junzhou Huang}.} \bibinfo{year}{2019}\natexlab{}.
\newblock \showarticletitle{Semi-supervised graph classification: A hierarchical graph perspective}. In \bibinfo{booktitle}{\emph{The World Wide Web Conference}}. \bibinfo{pages}{972--982}.
\newblock


\bibitem[Li et~al\mbox{.}(2018)]%
        {li2017diffusion}
\bibfield{author}{\bibinfo{person}{Yaguang Li}, \bibinfo{person}{Rose Yu}, \bibinfo{person}{Cyrus Shahabi}, {and} \bibinfo{person}{Yan Liu}.} \bibinfo{year}{2018}\natexlab{}.
\newblock \showarticletitle{Diffusion Convolutional Recurrent Neural Network: Data-Driven Traffic Forecasting}. In \bibinfo{booktitle}{\emph{International Conference on Learning Representations}}.
\newblock


\bibitem[Lim et~al\mbox{.}(2021)]%
        {lim2021large}
\bibfield{author}{\bibinfo{person}{Derek Lim}, \bibinfo{person}{Felix Hohne}, \bibinfo{person}{Xiuyu Li}, \bibinfo{person}{Sijia~Linda Huang}, \bibinfo{person}{Vaishnavi Gupta}, \bibinfo{person}{Omkar Bhalerao}, {and} \bibinfo{person}{Ser~Nam Lim}.} \bibinfo{year}{2021}\natexlab{}.
\newblock \showarticletitle{Large scale learning on non-homophilous graphs: New benchmarks and strong simple methods}.
\newblock \bibinfo{journal}{\emph{Advances in Neural Information Processing Systems}}  \bibinfo{volume}{34} (\bibinfo{year}{2021}), \bibinfo{pages}{20887--20902}.
\newblock


\bibitem[Luan et~al\mbox{.}(2022)]%
        {abs-2210-07606}
\bibfield{author}{\bibinfo{person}{Sitao Luan}, \bibinfo{person}{Chenqing Hua}, \bibinfo{person}{Qincheng Lu}, \bibinfo{person}{Jiaqi Zhu}, \bibinfo{person}{Mingde Zhao}, \bibinfo{person}{Shuyuan Zhang}, \bibinfo{person}{Xiao-Wen Chang}, {and} \bibinfo{person}{Doina Precup}.} \bibinfo{year}{2022}\natexlab{}.
\newblock \showarticletitle{Revisiting Heterophily For Graph Neural Networks}. In \bibinfo{booktitle}{\emph{Advances in Neural Information Processing Systems}}.
\newblock


\bibitem[Mattei and Frellsen(2019)]%
        {mattei2019miwae}
\bibfield{author}{\bibinfo{person}{Pierre-Alexandre Mattei} {and} \bibinfo{person}{Jes Frellsen}.} \bibinfo{year}{2019}\natexlab{}.
\newblock \showarticletitle{MIWAE: Deep generative modelling and imputation of incomplete data sets}. In \bibinfo{booktitle}{\emph{International conference on machine learning}}. PMLR, \bibinfo{pages}{4413--4423}.
\newblock


\bibitem[Mazumder et~al\mbox{.}(2010)]%
        {mazumder2010spectral}
\bibfield{author}{\bibinfo{person}{Rahul Mazumder}, \bibinfo{person}{Trevor Hastie}, {and} \bibinfo{person}{Robert Tibshirani}.} \bibinfo{year}{2010}\natexlab{}.
\newblock \showarticletitle{Spectral regularization algorithms for learning large incomplete matrices}.
\newblock \bibinfo{journal}{\emph{The Journal of Machine Learning Research}}  \bibinfo{volume}{11} (\bibinfo{year}{2010}), \bibinfo{pages}{2287--2322}.
\newblock


\bibitem[Min et~al\mbox{.}(2020)]%
        {min2020scattering}
\bibfield{author}{\bibinfo{person}{Yimeng Min}, \bibinfo{person}{Frederik Wenkel}, {and} \bibinfo{person}{Guy Wolf}.} \bibinfo{year}{2020}\natexlab{}.
\newblock \showarticletitle{Scattering gcn: Overcoming oversmoothness in graph convolutional networks}.
\newblock \bibinfo{journal}{\emph{Advances in Neural Information Processing Systems}}  \bibinfo{volume}{33} (\bibinfo{year}{2020}), \bibinfo{pages}{14498--14508}.
\newblock


\bibitem[Mistler and Enders(2017)]%
        {mistler2017comparison}
\bibfield{author}{\bibinfo{person}{Stephen~A Mistler} {and} \bibinfo{person}{Craig~K Enders}.} \bibinfo{year}{2017}\natexlab{}.
\newblock \showarticletitle{A comparison of joint model and fully conditional specification imputation for multilevel missing data}.
\newblock \bibinfo{journal}{\emph{Journal of Educational and Behavioral Statistics}} \bibinfo{volume}{42}, \bibinfo{number}{4} (\bibinfo{year}{2017}), \bibinfo{pages}{432--466}.
\newblock


\bibitem[Mohan et~al\mbox{.}(2013)]%
        {mohan2013graphical}
\bibfield{author}{\bibinfo{person}{Karthika Mohan}, \bibinfo{person}{Judea Pearl}, {and} \bibinfo{person}{Jin Tian}.} \bibinfo{year}{2013}\natexlab{}.
\newblock \showarticletitle{Graphical models for inference with missing data}.
\newblock \bibinfo{journal}{\emph{Advances in neural information processing systems}}  \bibinfo{volume}{26} (\bibinfo{year}{2013}).
\newblock


\bibitem[Morris et~al\mbox{.}(2016)]%
        {morris2016faster}
\bibfield{author}{\bibinfo{person}{Christopher Morris}, \bibinfo{person}{Nils~M Kriege}, \bibinfo{person}{Kristian Kersting}, {and} \bibinfo{person}{Petra Mutzel}.} \bibinfo{year}{2016}\natexlab{}.
\newblock \showarticletitle{Faster kernels for graphs with continuous attributes via hashing}. In \bibinfo{booktitle}{\emph{2016 IEEE 16th International Conference on Data Mining (ICDM)}}. IEEE, \bibinfo{pages}{1095--1100}.
\newblock


\bibitem[Muzellec et~al\mbox{.}(2020)]%
        {muzellec2020missing}
\bibfield{author}{\bibinfo{person}{Boris Muzellec}, \bibinfo{person}{Julie Josse}, \bibinfo{person}{Claire Boyer}, {and} \bibinfo{person}{Marco Cuturi}.} \bibinfo{year}{2020}\natexlab{}.
\newblock \showarticletitle{Missing data imputation using optimal transport}. In \bibinfo{booktitle}{\emph{International Conference on Machine Learning}}. PMLR, \bibinfo{pages}{7130--7140}.
\newblock


\bibitem[Orsini et~al\mbox{.}(2015)]%
        {orsini2015graph}
\bibfield{author}{\bibinfo{person}{Francesco Orsini}, \bibinfo{person}{Paolo Frasconi}, {and} \bibinfo{person}{Luc De~Raedt}.} \bibinfo{year}{2015}\natexlab{}.
\newblock \showarticletitle{Graph invariant kernels}. In \bibinfo{booktitle}{\emph{Twenty-Fourth International Joint Conference on Artificial Intelligence}}.
\newblock


\bibitem[Park et~al\mbox{.}(2019)]%
        {park2019symmetric}
\bibfield{author}{\bibinfo{person}{Jiwoong Park}, \bibinfo{person}{Minsik Lee}, \bibinfo{person}{Hyung~Jin Chang}, \bibinfo{person}{Kyuewang Lee}, {and} \bibinfo{person}{Jin~Young Choi}.} \bibinfo{year}{2019}\natexlab{}.
\newblock \showarticletitle{Symmetric graph convolutional autoencoder for unsupervised graph representation learning}. In \bibinfo{booktitle}{\emph{Proceedings of the IEEE/CVF International Conference on Computer Vision}}. \bibinfo{pages}{6519--6528}.
\newblock


\bibitem[Rabin and Fishelov(2017)]%
        {rabin2017missing}
\bibfield{author}{\bibinfo{person}{Neta Rabin} {and} \bibinfo{person}{Dalia Fishelov}.} \bibinfo{year}{2017}\natexlab{}.
\newblock \showarticletitle{Missing data completion using diffusion maps and laplacian pyramids}. In \bibinfo{booktitle}{\emph{International Conference on Computational Science and Its Applications}}. Springer, \bibinfo{pages}{284--297}.
\newblock


\bibitem[Rabin and Fishelov(2019)]%
        {rabin2019two}
\bibfield{author}{\bibinfo{person}{Neta Rabin} {and} \bibinfo{person}{Dalia Fishelov}.} \bibinfo{year}{2019}\natexlab{}.
\newblock \showarticletitle{Two directional Laplacian pyramids with application to data imputation}.
\newblock \bibinfo{journal}{\emph{Advances in Computational Mathematics}} \bibinfo{volume}{45}, \bibinfo{number}{4} (\bibinfo{year}{2019}), \bibinfo{pages}{2123--2146}.
\newblock


\bibitem[Rossi et~al\mbox{.}(2021)]%
        {rossi2021unreasonable}
\bibfield{author}{\bibinfo{person}{Emanuele Rossi}, \bibinfo{person}{Henry Kenlay}, \bibinfo{person}{Maria~I Gorinova}, \bibinfo{person}{Benjamin~Paul Chamberlain}, \bibinfo{person}{Xiaowen Dong}, {and} \bibinfo{person}{Michael Bronstein}.} \bibinfo{year}{2021}\natexlab{}.
\newblock \showarticletitle{On the Unreasonable Effectiveness of Feature propagation in Learning on Graphs with Missing Node Features}.
\newblock \bibinfo{journal}{\emph{arXiv preprint arXiv:2111.12128}} (\bibinfo{year}{2021}).
\newblock


\bibitem[Rubin(1976)]%
        {rubin1976inference}
\bibfield{author}{\bibinfo{person}{Donald~B Rubin}.} \bibinfo{year}{1976}\natexlab{}.
\newblock \showarticletitle{Inference and missing data}.
\newblock \bibinfo{journal}{\emph{Biometrika}} \bibinfo{volume}{63}, \bibinfo{number}{3} (\bibinfo{year}{1976}), \bibinfo{pages}{581--592}.
\newblock


\bibitem[Sportisse et~al\mbox{.}(2020)]%
        {SportisseBJ20}
\bibfield{author}{\bibinfo{person}{Aude Sportisse}, \bibinfo{person}{Claire Boyer}, {and} \bibinfo{person}{Julie Josse}.} \bibinfo{year}{2020}\natexlab{}.
\newblock \showarticletitle{Estimation and imputation in probabilistic principal component analysis with missing not at random data}.
\newblock \bibinfo{journal}{\emph{Advances in Neural Information Processing Systems}}  \bibinfo{volume}{33} (\bibinfo{year}{2020}), \bibinfo{pages}{7067--7077}.
\newblock


\bibitem[Stekhoven and B{\"u}hlmann(2012)]%
        {stekhoven2012missforest}
\bibfield{author}{\bibinfo{person}{Daniel~J Stekhoven} {and} \bibinfo{person}{Peter B{\"u}hlmann}.} \bibinfo{year}{2012}\natexlab{}.
\newblock \showarticletitle{MissForest—non-parametric missing value imputation for mixed-type data}.
\newblock \bibinfo{journal}{\emph{Bioinformatics}} (\bibinfo{year}{2012}).
\newblock


\bibitem[Taguchi et~al\mbox{.}(2021)]%
        {taguchi2021graph}
\bibfield{author}{\bibinfo{person}{Hibiki Taguchi}, \bibinfo{person}{Xin Liu}, {and} \bibinfo{person}{Tsuyoshi Murata}.} \bibinfo{year}{2021}\natexlab{}.
\newblock \showarticletitle{Graph convolutional networks for graphs containing missing features}.
\newblock \bibinfo{journal}{\emph{Future Generation Computer Systems}}  \bibinfo{volume}{117} (\bibinfo{year}{2021}), \bibinfo{pages}{155--168}.
\newblock


\bibitem[Troyanskaya et~al\mbox{.}(2001a)]%
        {troyanskaya2001missing}
\bibfield{author}{\bibinfo{person}{Olga Troyanskaya}, \bibinfo{person}{Michael Cantor}, \bibinfo{person}{Gavin Sherlock}, \bibinfo{person}{Pat Brown}, \bibinfo{person}{Trevor Hastie}, \bibinfo{person}{Robert Tibshirani}, \bibinfo{person}{David Botstein}, {and} \bibinfo{person}{Russ~B Altman}.} \bibinfo{year}{2001}\natexlab{a}.
\newblock \showarticletitle{Missing value estimation methods for DNA microarrays}.
\newblock \bibinfo{journal}{\emph{Bioinformatics}} \bibinfo{volume}{17}, \bibinfo{number}{6} (\bibinfo{year}{2001}), \bibinfo{pages}{520--525}.
\newblock


\bibitem[Troyanskaya et~al\mbox{.}(2001b)]%
        {TroyanskayaCSBHTBA01}
\bibfield{author}{\bibinfo{person}{Olga~G. Troyanskaya}, \bibinfo{person}{Michael~N. Cantor}, \bibinfo{person}{Gavin Sherlock}, \bibinfo{person}{Patrick~O. Brown}, \bibinfo{person}{Trevor Hastie}, \bibinfo{person}{Robert Tibshirani}, \bibinfo{person}{David Botstein}, {and} \bibinfo{person}{Russ~B. Altman}.} \bibinfo{year}{2001}\natexlab{b}.
\newblock \showarticletitle{Missing value estimation methods for {DNA} microarrays}.
\newblock \bibinfo{journal}{\emph{Bioinform.}} \bibinfo{volume}{17}, \bibinfo{number}{6} (\bibinfo{year}{2001}), \bibinfo{pages}{520--525}.
\newblock


\bibitem[Van~Buuren(2007)]%
        {van2007multiple}
\bibfield{author}{\bibinfo{person}{Stef Van~Buuren}.} \bibinfo{year}{2007}\natexlab{}.
\newblock \showarticletitle{Multiple imputation of discrete and continuous data by fully conditional specification}.
\newblock \bibinfo{journal}{\emph{Statistical methods in medical research}} \bibinfo{volume}{16}, \bibinfo{number}{3} (\bibinfo{year}{2007}), \bibinfo{pages}{219--242}.
\newblock


\bibitem[Van~Buuren and Groothuis-Oudshoorn(2011)]%
        {van2011mice}
\bibfield{author}{\bibinfo{person}{Stef Van~Buuren} {and} \bibinfo{person}{Karin Groothuis-Oudshoorn}.} \bibinfo{year}{2011}\natexlab{}.
\newblock \showarticletitle{mice: Multivariate imputation by chained equations in R}.
\newblock \bibinfo{journal}{\emph{Journal of statistical software}}  \bibinfo{volume}{45} (\bibinfo{year}{2011}), \bibinfo{pages}{1--67}.
\newblock


\bibitem[Wu et~al\mbox{.}(2019)]%
        {wu2019simplifying}
\bibfield{author}{\bibinfo{person}{Felix Wu}, \bibinfo{person}{Amauri Souza}, \bibinfo{person}{Tianyi Zhang}, \bibinfo{person}{Christopher Fifty}, \bibinfo{person}{Tao Yu}, {and} \bibinfo{person}{Kilian Weinberger}.} \bibinfo{year}{2019}\natexlab{}.
\newblock \showarticletitle{Simplifying graph convolutional networks}. In \bibinfo{booktitle}{\emph{International conference on machine learning}}. PMLR, \bibinfo{pages}{6861--6871}.
\newblock


\bibitem[Wu et~al\mbox{.}(2023)]%
        {wu2023differentiable}
\bibfield{author}{\bibinfo{person}{Yangyang Wu}, \bibinfo{person}{Jun Wang}, \bibinfo{person}{Xiaoye Miao}, \bibinfo{person}{Wenjia Wang}, {and} \bibinfo{person}{Jianwei Yin}.} \bibinfo{year}{2023}\natexlab{}.
\newblock \showarticletitle{Differentiable and scalable generative adversarial models for data imputation}.
\newblock \bibinfo{journal}{\emph{IEEE Transactions on Knowledge and Data Engineering}} (\bibinfo{year}{2023}).
\newblock


\bibitem[Wu et~al\mbox{.}(2021)]%
        {wu2021inductive}
\bibfield{author}{\bibinfo{person}{Yuankai Wu}, \bibinfo{person}{Dingyi Zhuang}, \bibinfo{person}{Aurelie Labbe}, {and} \bibinfo{person}{Lijun Sun}.} \bibinfo{year}{2021}\natexlab{}.
\newblock \showarticletitle{Inductive Graph Neural Networks for Spatiotemporal Kriging}. In \bibinfo{booktitle}{\emph{Proceedings of the AAAI Conference on Artificial Intelligence}}, Vol.~\bibinfo{volume}{35}. \bibinfo{pages}{4478--4485}.
\newblock


\bibitem[Yoon et~al\mbox{.}(2018)]%
        {yoon2018gain}
\bibfield{author}{\bibinfo{person}{Jinsung Yoon}, \bibinfo{person}{James Jordon}, {and} \bibinfo{person}{Mihaela Schaar}.} \bibinfo{year}{2018}\natexlab{}.
\newblock \showarticletitle{Gain: Missing data imputation using generative adversarial nets}. In \bibinfo{booktitle}{\emph{International conference on machine learning}}. PMLR, \bibinfo{pages}{5689--5698}.
\newblock


\bibitem[You et~al\mbox{.}(2020)]%
        {you2020handling}
\bibfield{author}{\bibinfo{person}{Jiaxuan You}, \bibinfo{person}{Xiaobai Ma}, \bibinfo{person}{Yi Ding}, \bibinfo{person}{Mykel~J Kochenderfer}, {and} \bibinfo{person}{Jure Leskovec}.} \bibinfo{year}{2020}\natexlab{}.
\newblock \showarticletitle{Handling missing data with graph representation learning}.
\newblock \bibinfo{journal}{\emph{Advances in Neural Information Processing Systems}}  \bibinfo{volume}{33} (\bibinfo{year}{2020}), \bibinfo{pages}{19075--19087}.
\newblock


\bibitem[Zheng et~al\mbox{.}(2022)]%
        {zheng2022graph}
\bibfield{author}{\bibinfo{person}{Xin Zheng}, \bibinfo{person}{Yixin Liu}, \bibinfo{person}{Shirui Pan}, \bibinfo{person}{Miao Zhang}, \bibinfo{person}{Di Jin}, {and} \bibinfo{person}{Philip~S Yu}.} \bibinfo{year}{2022}\natexlab{}.
\newblock \showarticletitle{Graph neural networks for graphs with heterophily: A survey}.
\newblock \bibinfo{journal}{\emph{arXiv preprint arXiv:2202.07082}} (\bibinfo{year}{2022}).
\newblock


\bibitem[Zhou et~al\mbox{.}(2021)]%
        {zhou2021dirichlet}
\bibfield{author}{\bibinfo{person}{Kaixiong Zhou}, \bibinfo{person}{Xiao Huang}, \bibinfo{person}{Daochen Zha}, \bibinfo{person}{Rui Chen}, \bibinfo{person}{Li Li}, \bibinfo{person}{Soo-Hyun Choi}, {and} \bibinfo{person}{Xia Hu}.} \bibinfo{year}{2021}\natexlab{}.
\newblock \showarticletitle{Dirichlet energy constrained learning for deep graph neural networks}.
\newblock \bibinfo{journal}{\emph{Advances in Neural Information Processing Systems}}  \bibinfo{volume}{34} (\bibinfo{year}{2021}).
\newblock


\bibitem[Zhu et~al\mbox{.}(2020)]%
        {zhu2020beyond}
\bibfield{author}{\bibinfo{person}{Jiong Zhu}, \bibinfo{person}{Yujun Yan}, \bibinfo{person}{Lingxiao Zhao}, \bibinfo{person}{Mark Heimann}, \bibinfo{person}{Leman Akoglu}, {and} \bibinfo{person}{Danai Koutra}.} \bibinfo{year}{2020}\natexlab{}.
\newblock \showarticletitle{Beyond homophily in graph neural networks: Current limitations and effective designs}.
\newblock \bibinfo{journal}{\emph{Advances in Neural Information Processing Systems}}  \bibinfo{volume}{33} (\bibinfo{year}{2020}), \bibinfo{pages}{7793--7804}.
\newblock


\end{thebibliography}

\end{document}